\documentclass[10pt,twocolumn,twoside]{IEEEtran}
\usepackage[pdftex]{graphicx}
\graphicspath{{./fig/}{./fig/syn/}{./fig/real/}}
\DeclareGraphicsExtensions{.pdf,.jpeg,.png}
\usepackage[cmex10]{amsmath}
\usepackage{amssymb}
\usepackage{amsthm}
\usepackage{amsfonts}
\usepackage{epstopdf}
\usepackage{multirow}
\usepackage{algorithm}
\usepackage{algorithmic}
\usepackage{subfigure}
\usepackage{url}
\usepackage{color}

\usepackage{mathtools}

\begin{document}
\title{Learning Laplacian Matrix in Smooth Graph Signal Representations}
\author{Xiaowen~Dong,
        Dorina~Thanou,
        Pascal~Frossard,
        and Pierre~Vandergheynst
\thanks{X. Dong is with the MIT Media Lab, Cambridge, MA, United States (e-mail: xdong@mit.edu).}
\thanks{D. Thanou, P. Frossard and P. Vandergheynst are with the Signal Processing Laboratories (LTS4/LTS2), \'{E}cole Polytechnique F\'{e}d\'{e}rale de Lausanne (EPFL), Lausanne, Switzerland (e-mail: \{dorina.thanou, pascal.frossard, pierre.vandergheynst\}@epfl.ch).}
\thanks{X. Dong is supported by a Swiss National Science Foundation Mobility fellowship. This work was mostly done while X. Dong was at EPFL. The work has been partially funded by the Hasler Foundation, Switzerland, under the project LOGAN, and the Swiss National Science Foundation under Grant 200021\_135493.}}
\maketitle

\begin{abstract}
The construction of a meaningful graph plays a crucial role in the success of many graph-based representations and algorithms for handling structured data, especially in the emerging field of graph signal processing. However, a meaningful graph is not always readily available from the data, nor easy to define depending on the application domain. In particular, it is often desirable in graph signal processing applications that a graph is chosen such that the data admit certain regularity or smoothness on the graph. In this paper, we address the problem of learning graph Laplacians, which is equivalent to learning graph topologies, such that the input data form graph signals with smooth variations on the resulting topology. To this end, we adopt a factor analysis model for the graph signals and impose a Gaussian probabilistic prior on the latent variables that control these signals. We show that the Gaussian prior leads to an efficient representation that favors the smoothness property of the graph signals. We then propose an algorithm for learning graphs that enforces such property and is based on minimizing the variations of the signals on the learned graph. Experiments on both synthetic and real world data demonstrate that the proposed graph learning framework can efficiently infer meaningful graph topologies from signal observations under the smoothness prior.
\end{abstract}


\IEEEpeerreviewmaketitle

\section{Introduction}
\label{sec:intro}
\IEEEPARstart{M}{odern} data processing tasks often manipulate structured data, where signal values are defined on the vertex set of a weighted and undirected graph. We refer to such data as graph signals, where the vertices of the graph represent the entities and the edge weights reflect the pairwise relationships between these entities. The signal values associated with the vertices carry the information of interest in observations or physical measurements. Numerous examples can be found in real world applications, such as temperatures within a geographical area, transportation capacities at hubs in a transportation network, or human behaviors in a social network. Such data representations have led to the emerging field of graph signal processing \cite{Shuman13a,Sandryhaila13}, which studies the representation, approximation and processing of such structured signals. 
Currently, most of the research effort in this field has been devoted to the analysis and processing of the signals, which are defined on a graph that is a priori known or naturally chosen from the application domain, e.g., geographical or social friendship graphs. However, these natural choices of graphs may not necessarily describe well the intrinsic relationships between the entities in the data. Furthermore, a natural graph might not be easy to define at all in some applications. When a good graph is not readily available, it is desirable to learn the graph topology from the observed data such that it captures well the intrinsic relationships between the entities, and permits effective data processing and analysis. This is exactly the main objective of the present paper.

Generally speaking, learning graphs from data samples is an ill-posed problem, and there might be many solutions to associate a structure to the data. We therefore need to define meaningful models for computing the topology such that the relationships between the signals and the graph topology satisfy these pre-defined models. This allows us to define meaningful criteria to describe or estimate structures between the data samples. In this work, we consider the large class of signals that are smooth on a graph structure. A graph signal is smooth if the signal values associated with the two end vertices of edges with large weights in the graph tend to be similar. Such property is often observed in real world graph signals. For instance, consider a geographical graph where the vertices represent different locations and the edges capture their physical proximities. If we define the signal values as temperature records observed in these locations, then the values at close-by vertices tend to be similar in regions without significant terrain variations. Another example would be a social network graph where the vertices and edges respectively represent people and the friendships between them, and signals defined as personal interests. It is often the case that friends, represented by connected vertices, share common interests. In both examples, data have smooth graph signal representations as long as the graphs are appropriately chosen. Smooth signal models on graphs have been used in many problems involving regularization on graphs \cite{Smola03}, with applications to signal denoising \cite{Shuman13a}, classification \cite{Zhou04} or multi-view clustering \cite{Dong12} to name a few.

\begin{figure}[t]
	\begin{center}
		\begin{tabular}{cc}
			~\includegraphics[width=0.45\textwidth]{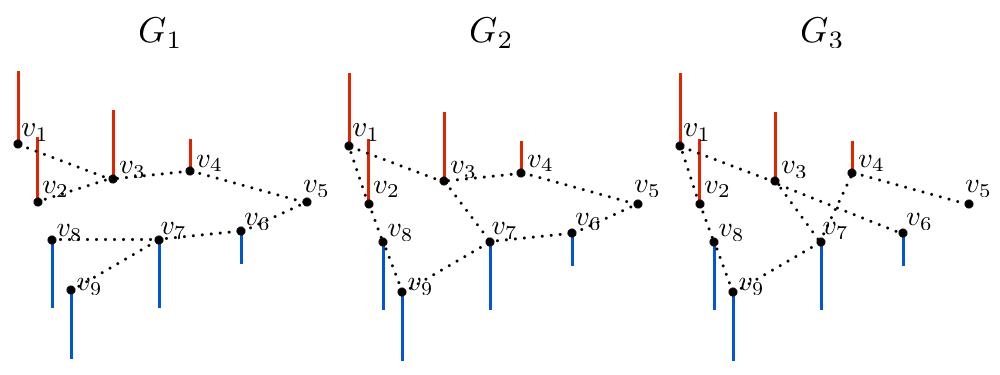}~ 
		\end{tabular}
	\end{center}
	\vspace{-0.2cm}
	\caption{A given signal can potentially live on different graphs but only one leads to a smooth graph signal (figure inspired by \cite{Shuman13a}). The red bars pointing upwards and the blue bars pointing downwards represent positive and negative signal values, respectively. The length of the bars reflects the magnitude of the values. In this illustration, while all the graphs are valid a priori, only choosing graph $G_1$ would favor the smoothness property of the resulting graph signal.}
	\label{fig:graphlearning}
\end{figure}

To further illustrate the interplay between the graph and the data, we consider a signal given by a set of unordered scalar values, which can potentially be defined on three different graphs $G_1$, $G_2$ and $G_3$, leading to three graph signals shown in Fig.~\ref{fig:graphlearning}. Without any assumption on the properties of the graph signal, the three candidate graphs are all valid choices. However, if we assume that the signal is smooth on the underlying graph, then $G_1$ is obviously a more reasonable choice than $G_2$ and $G_3$. Our objective is exactly to learn a graph similar to $G_1$ when the signal is expected to be smooth on an ``unknown'' graph topology. Formally speaking, we consider in this paper the following problem:
\textit{Given a set of potentially noisy data $X=\{x_i\}_{i=1}^p$ ($x_i \in \mathbb{R}^n$), we would like to infer an optimal weighted and undirected graph $G$ of $n$ vertices, namely, its edges and the associated weights, such that the elements of $X$ form smooth signals on $G$.}

In this paper, we build on our previous work in \cite{Dong15} and propose to define the relationships between signals and graphs by revisiting the representation learning theory \cite{Bengio13} studied in the traditional signal processing setting. Specifically, we consider a factor analysis model used in the theory of representation learning, where we assume that the observed signals are controlled by a set of unobserved latent variables with some given probabilistic prior. 
We generalize this model to graph signals, where we assume that the observed graph signals can be represented in terms of a set of latent variables. The difference from the traditional analysis model is that the transformation from the latent variables to the observed signals involves information about the topology of the graph. As a result, we can define joint properties (or a joint model) between the signals and the graph, such that the signal representations are consistent with given statistical priors on the latent variables. Specifically, by imposing a Gaussian prior on the latent variables in our generalized factor analysis model, we obtain a Principal Component Analysis (PCA)-like representation for the graph signals, which turns out to be a smooth signal representation on graphs. The relationship between the Gaussian latent variables and the signal observations is based on the graph Laplacian matrix, which uniquely characterizes the graph topology and has been widely connected in the graph signal processing literature with the notion of smoothness on the graph. Hence, learning a graph from data becomes equivalent to learning its graph Laplacian matrix, which is central to many key concepts in graph signal processing.

Building on the above model, we design a new algorithm for learning a valid graph Laplacian operator from data samples, such that the graph signal representation is smooth and consistent with the Gaussian prior on the latent variables. Specifically, given potentially noisy signal observations, our algorithm iterates between the learning of a smoothed version of the original signal measurements and the learning of a graph topology such that the variations of the smooth signals on the learned graph are minimized upon convergence. We test our graph learning algorithm on several synthetic and real world experiments, where we show that it is able to efficiently infer the topology of the groundtruth graphs. It further achieves competitive performance with a state-of-the-art structural learning algorithm, which is closely related to the idea of sparse inverse covariance estimation for Gaussian graphical models \cite{Lake10}. More importantly, our results demonstrate the meaningfulness and efficiency of a new learning framework from the graph signal processing perspective.



Our novel approach is one of the first rigorous frameworks in graph signal processing that connects the learning of the graph topology to the representation model of the graph signals and their properties. By extending the classical and generic factor analysis model to graph settings, our framework provides new insights into the understanding of the interactions between signals and graphs. 
We believe that our new framework can benefit numerous emerging real world applications, such as the analysis of transportation, biomedical, and social networks, where it is critical to infer hidden relationships between data entities.


\section{Related work}
\label{sec:literature}
Graph signal processing, as an emerging research field, has been attracting an increasing amount of interest from the signal processing community. Most of the research effort so far has been devoted to the efficient representation and processing of signals defined on a given graph. Representative works include studies on the generalizations of the Fourier transform \cite{Shuman12,Zhu12}, the wavelet transform \cite{Coifman06,Gavish10,Hammond11,Ram11,Narang12,Shuman13c}, dictionary learning methods \cite{Zhang12,Thanou14}, and time-frequency analysis \cite{Shuman13d,Agaskar13} on graphs, with applications to image processing \cite{Shen10}, brain network classification \cite{Hu13b}, mobility inference \cite{Dong13}, and community detection \cite{Tremblay14} to name a few. In most of these works, the irregular domain of the graph is captured through the graph Laplacian operator, whose importance can be summarized as follows. First, from a graph signal processing perspective, a Laplacian matrix represents a valid graph topology $G$ (i.e., with non-negative edge weights and without self-loops as usually considered in the literature), and supports the definition of graph signals. Furthermore, the Laplacian matrix enables the generalization of the notion of frequency and Fourier transform for graph signals \cite{Hammond11}, which is the building block of many techniques that have been recently developed in the field of graph signal processing. Therefore, the Laplacian matrix is directly related to the processing of graph signals. Second, the graph Laplacian is an operator central to the definition of various kernels on graphs via its spectral decomposition. More specifically, Smola et al. has shown in \cite{Smola03} that graph Laplacian can be used to define various kernels related to regularization, diffusion, and random walk processes associated with graphs. 
As a result, the Laplacian matrix permits to process real world graph signals that represent complex behaviors associated with kernels on graphs. Third, the graph Laplacian, under some conditions, converges to the intrinsic Laplace-Beltrami operator on a Riemannian manifold \cite{Hein05,Singer06,Belkin08}, therefore it may well capture the underlying manifold structure of the observed signals hence benefit the processing tasks. Finally, from an application point of view, Zhang et al. has shown that, when a signal follows a Gaussian Markov Random Field (GMRF) model with the precision matrix (inverse covariance matrix) defined as the graph Laplacian, the eigenvectors of the Laplacian optimally decorrelate the signal and can thus be used for efficient compression \cite{Zhang13}. These results show that the knowledge of a valid Laplacian matrix is indeed beneficial to graph-based signal and image processing in real world applications.

While many algorithms are built on the graph Laplacian, they mainly assume that the later is either given a priori, or chosen naturally from the application domain. Relatively less research effort has been devoted to the construction or learning of the graph topologies. A regression framework is proposed in \cite{Hu13a} to learn a graph Laplacian matrix based on a fitness metric between the signals and the graph that essentially evaluates the smoothness of the signals on the graph. A similar fitness metric has also been used in \cite{Daitch09} to learn a valid graph topology (the adjacency matrix). The main difference between these two works and the method proposed in this paper is that, instead of using a fitness metric, our work relates the graph learning process and the properties of the graph signals in a statistical manner that helps in understanding the data properties. Moreover, the factor analysis model adopted in our new framework implies that the graph signals are generated by a simple linear statistical model, which can be revealed through the learning process. 
Two other graph learning works have been described in \cite{Leonardi11} and \cite{Leonardi13}, where the authors have proposed to use, respectively, correlations between wavelet coefficients of the time series of brain signals, and Principal Component Analysis applied to correlation matrices, to estimate functional connectivities of distinct brain regions. Although these can be considered as methods that learn an activity graph between the brain regions, they are essentially similar to traditional approaches that are based on pairwise correlation analysis rather than considering the global behavior of the signals on the graph.
Finally, the problem of constructing a meaningful graph has been studied implicitly in \cite{Argyriou,Dai2007,ShiJebECML2010} in the context of multiple kernel learning. The first two works aim at finding an optimal convex combination of Laplacians of some initial graphs that are constructed a priori using some known features. One main difference with our approach is that, in these two papers the basic graphs are constructed using distance metrics and additional features, and a label function is learned by enforcing its smoothness over all the basic graphs through a convex combination. On the other hand, in our work, we do not assume any prior knowledge or additional information to construct such basic graphs and the optimal graph is learned by only using a set of signal observations. The authors in \cite{ShiJebECML2010} learn an optimal transformation of the spectrum of a predefined graph Laplacian while fixing its eigenvector space (such a graph is again computed using additional features). In comparison, our work aims at learning both the spectrum and eigenspace (hence the Laplacian matrix) from the mere signal measurements.


In the meantime, there is a large amount of work from the machine learning community that aims at solving similar learning problems. In particular, one topic in the research of learning graphical models is to estimate from the observed data an inverse covariance matrix (precision matrix) for Gaussian graphical models \cite{Meinshausen06,Yuan07,Banerjee08,Friedman08,Rothman08,Ravikumar08,Hsieh11}, especially in the case when the number of observations is smaller than the sample dimension and the sample covariance becomes singular. Another problem consists in inferring the graph structure for discrete Markov Random Fields \cite{Loh13}. It is known that in case of a Gaussian graphical model, there is an exact correspondence between the location of the non-zero entries in the precision matrix and the existence of partial correlations between the random variables \cite{Rue05}. In this case, a maximum-likelihood estimator turns out to be the solution to a log-determinant program. The estimated precision matrix is therefore considered to carry information about the partial correlations between the random variables. We would however like to emphasize the difference between this type of approaches and our framework by making the following points. First, it is important to notice that in our framework we aim at learning a valid graph Laplacian, which is not the case in the above mentioned works. Indeed, in those cases the learned precision matrix is a full-rank matrix that usually (i) has both positive and negative off-diagonal entries reflecting both positive and negative correlations, and (ii) does not have rows summed up to zero. Non-positive off-diagonal entries and zero row-sum are however necessary constraints to define a graph Laplacian. As a result, the learning of the precision matrix in those methods is not directly linked to the interpretation of global properties of the input graph signals but rather reflects the partial correlations between the random variables that control the observations. Second, it is not straightforward to project the learned precision matrix in those approaches onto the convex set of the constraints that are needed in order to transform it into a valid Laplacian matrix. Therefore, we rather propose in this work to learn a Laplacian matrix directly through the optimization problem. Finally, the work in \cite{Lake10} learns a valid graph topology (the adjacency matrix) with an optimization problem that is very similar to the classical problem of sparse inverse covariance estimation, but with a regularized full-rank Laplacian matrix. In the specific case of an infinite a priori feature variance in the problem formulation, this approach looks similar to ours but with an explicit sparsity-promoting term for the adjacency matrix.
To summarize, an essential difference between our method and the above-mentioned machine learning approaches is that, most of these methods do not learn a valid graph Laplacian matrix; they mainly focus on the pairwise correlation between random variables (including the work in \cite{Lake10}), but do not explore the link between the signal model (i.e., global smoothness) and the graph topology. This is exactly the central focus of this paper, whose objective is to learn a representation that permits effective analysis of the graph signals.

\section{Graph signal representation via factor analysis}
\label{sec:laplacian}

\subsection{Smooth graph signals}
We cast the problem of learning graph topology as a problem of learning the so-called graph Laplacian matrix as it uniquely characterizes the graph. Consider a weighted and undirected graph $G=\{V,E,\omega\}$ of $n$ vertices, where $V$ and $E$ represent the vertex and edge sets, respectively, and $\omega$ gives a positive weight for each edge in $E$. A graph signal can then be represented by a function $f: V \rightarrow \mathbb{R}^n$ that assigns a scalar value to each vertex. The unnormalized (or combinatorial) graph Laplacian matrix $L$, which is an $n$ by $n$ matrix, is defined as:
\begin{equation}
L=D-W,
\label{eq:laplacian}
\end{equation}
where $W$ is the possibly weighted adjacency matrix of the graph and $D$ is the degree matrix that contains the degrees of the vertices along the diagonal. Since $L$ is a real and symmetric matrix, it has a complete set of orthonormal eigenvectors and associated eigenvalues. In particular, it can be written as:
\begin{equation}
L = \chi \Lambda \chi^T,
\label{eq:eigendecomp}
\end{equation}
where $\chi$ is the eigenvector matrix that contains the eigenvectors as columns, and $\Lambda$ is the diagonal eigenvalue matrix where the associated eigenvalues are sorted in increasing order. The smallest eigenvalue is 0 with a multiplicity equal to the number of connected components of the graph $G$ \cite{Chung97}. Therefore, if $G$ has $c$ connected components, then the rank of $\Lambda$ is $n-c$. The Laplacian matrix $L$ enables, via its spectral decomposition in Eq.~(\ref{eq:eigendecomp}), the generalization of the notion of frequency and Fourier transform for graph signals \cite{Hammond11}. This leads to the extension of classical signal processing tools on graphs (see \cite{Shuman13a} for a detailed survey of these tools).

Recall that we consider in this paper the class of smooth signals on graphs. Equipped with the Laplacian matrix, the smoothness of a graph signal $f$ on $G$ can be measured in terms of a quadratic form of the graph Laplacian \cite{Zhou04}:
\begin{equation}
f^T L f = \frac{1}{2} \sum_{i \sim j} w_{ij} \big( f(i)-f(j) \big)^2,
\label{eq:smoothness}
\end{equation}
where $w_{ij}$ represents the weight on the edge connecting two adjacent vertices $i$ and $j$, and $f(i)$ and $f(j)$ are the signal values associated with these two vertices. Intuitively, given that the weights are non-negative, Eq.~(\ref{eq:smoothness}) shows that a graph signal $f$ is considered to be smooth if strongly connected vertices (with a large weight on the edge between them) have similar values. In particular, the smaller the quadratic form in Eq.~(\ref{eq:smoothness}), the smoother the signal on the graph. Smooth graph signal models have been widely used in various learning problems such as regularization and semi-supervised learning on graphs \cite{Smola03,Zhou04}.

\subsection{Representation via factor analysis model}
We now propose a new representation model for smooth graph signals based on an extension of the traditional factor analysis model. The factor analysis model \cite{Bartholomew11,Basilevsky94} is a generic linear statistical model that tries to explain observations of a given dimension with a potentially smaller number of unobserved latent variables. Specifically, we consider the following representation model for the input graph signal:
\begin{equation}
x = \chi h + u_x + \epsilon,
\label{eq:fa_graph}
\end{equation}
where $x \in \mathbb{R}^n$ represents the observed graph signal, $h \in \mathbb{R}^n$ represents the latent variable that controls the graph signal $x$ through the eigenvector matrix $\chi$, and $u_x \in \mathbb{R}^n$ is the mean of $x$. As discussed in \cite{Tipping99}, we adopt an isotropic noise model, namely, we assume that $\epsilon$ follows a multivariate Gaussian distribution with mean zero and covariance $\sigma_\epsilon^2I_n$. The probability density function of $\epsilon$ is thus given by:
\begin{equation}
\epsilon \sim \mathcal{N}(0,\sigma_\epsilon^2I_n).
\label{eq:pn_graph}
\end{equation}

The model in Eq.~(\ref{eq:fa_graph}) can be considered as a generalization of the classical factor analysis model to graph signals, where the key is the choice of the representation matrix as the eigenvector matrix $\chi$ of the graph Laplacian $L$. This relates the graph signals to the latent variable. The motivation of this definition is twofolds. First, we seek a representation matrix that reflects the topology of the graph, such that the construction of the graph can be directly linked to the representation and properties of the graph signals. The eigenvector matrix of the graph Laplacian is a good candidate, since it provides a spectral embedding of the graph vertices that can subsequently be used for many graph-based analysis tasks such as graph partitioning \cite{Luxburg07}. Second, it can be interpreted as the graph Fourier basis for representing graph signals \cite{Hammond11}, which makes it a natural choice as the representation matrix in Eq.~(\ref{eq:fa_graph}).

As in the classical factor analysis model, we impose a Gaussian prior on the latent variable $h$ in Eq.~(\ref{eq:fa_graph}). Specifically, we assume that the latent variable $h$ follows a degenerate zero-mean multivariate Gaussian distribution with the precision matrix defined as the eigenvalue matrix $\Lambda$ of the graph Laplacian $L$, i.e.,
\begin{equation}
h \sim \mathcal{N}(0,\Lambda^\dagger),
\label{eq:ph_graph}
\end{equation}
where $\Lambda^\dagger$ is the Moore-Penrose pseudoinverse of $\Lambda$. Together with the definition of $\chi$ as the representation matrix and its Fourier interpretation, the above assumption on $h$ implies that the energy of the signal tends to lie mainly in the low frequency components\footnote{We note that the effect of the first (constant) eigenvector of the graph Laplacian is ignored as the corresponding entry in the diagonal of $\Lambda^\dagger$ is zero.}, hence it promotes the smoothness of the graph signal. 
Based on Eqs.~(\ref{eq:fa_graph}), (\ref{eq:pn_graph}) and (\ref{eq:ph_graph}), the conditional probability of $x$ given $h$, and the probability of $x$, are respectively given as:
\begin{eqnarray}
x|h \sim \mathcal{N}(\chi h + u_x, \sigma_\epsilon^2I_n), \\
\label{eq:pxcon_graph}
x \sim \mathcal{N}(u_x, L^\dagger + \sigma_\epsilon^2I_n),
\label{eq:px_graph}
\end{eqnarray}
where we have used in Eq.~(\ref{eq:px_graph}) the fact that the pseudoinverse of $L$, $L^\dagger$, admits the following eigendecomposition:
\begin{equation}
L^\dagger = \chi \Lambda^\dagger \chi^T.
\end{equation}

It can be seen from Eq.~(\ref{eq:px_graph}) that, in a noise-free scenario where $\sigma_\epsilon = 0$, $x$ also follows a degenerate multivariate Gaussian distribution. Moreover, $x$ can be seen as a GMRF with respect to the graph $G$, where the precision matrix is chosen to be the graph Laplacian $L$. Notice that the GMRF is a very generic model such that the precision matrix can be defined with much freedom, as long as its non-zero entries encode the partial correlations between the random variables, and as long as their locations correspond exactly to the edges in the graph \cite{Rue05}. In particular, the graph Laplacian $L$ is commonly adopted as the precision matrix of the GMRFs that model images with Gaussian probabilistic priors\footnote{Notice that this is not the case in most of the state-of-the-art approaches for inverse covariance estimation, where the inverse covariance or precision matrix needs to be a nonsingular and full-rank matrix that cannot be interpreted as a graph Laplacian.} \cite{Zhang13}. By defining the representation matrix in the factor analysis model as the eigenvector matrix $\chi$ and assuming that the latent variable follows a degenerate Gaussian distribution with the precision matrix $\Lambda$, we can therefore recover a GMRF model with a precision matrix equal to $L$ in a noise-free scenario.

In the presence of noise, we see from Eq.~(\ref{eq:px_graph}) that, under a Gaussian prior on the latent variable $h$ that follows Eq.~(\ref{eq:ph_graph}), the representation matrix $\chi$ is the eigenvector matrix of the covariance of $x$. Indeed, the covariance matrix $L^\dagger + \sigma_\epsilon^2I_n$ admits the following eigendecomposition:
\begin{equation}
L^\dagger + \sigma_\epsilon^2I_n = \chi (\Lambda^\dagger + \sigma_\epsilon^2I_n) \chi^T.
\end{equation}
This is analogous to the classical factor analysis model where the representation matrix spans the same subspace as the $k$ leading principal components of the covariance of $x$ under a Gaussian prior on $h$. It has been pointed out in \cite{Bengio13} that signal representation with the classical factor analysis model provides a probabilistic interpretation of the highly successful representation learned by the PCA, which was originally presented in \cite{Roweis97,Tipping99}. Because of the analogy mentioned above, the representation in Eq.~(\ref{eq:fa_graph}) can thus be considered as a PCA-like representation for the graph signal $x$. More importantly, we show in the next section that this promotes the smoothness properties for the signal on the graph, based on which we propose our novel learning framework.

\section{Graph Laplacian learning algorithm}
\label{sec:learning}
\subsection{Learning framework}
Given the observation $x$ and the multivariate Gaussian prior distribution of $h$ in Eq.~(\ref{eq:ph_graph}), we are now interested in a MAP estimate of $h$. Specifically, by applying Bayes' rule and assuming without loss of generality that $u_x = 0$, the MAP estimate of the latent variables $h$ can be written as follows \cite{Gribonval11}:
\begin{equation}
\begin{split}
h_\text{MAP}(x) \coloneqq &\arg \max_h p(h|x) \\
= &\arg \max_h p(x|h)p(h) \\
= &\arg \min_h \left(-\text{log}~p_E(x-\chi h) -\text{log}~p_H(h)\right),
\end{split}
\label{eq:map}
\end{equation}
where $p_E(\epsilon)=p_E(x-\chi h)$ and $p_H(h)$ represent the probability density function (p.d.f.) of the noise and the latent variable, respectively, and $p(h|x)$ is the conditional p.d.f. of $h$ given a realization of $x$.
Now, from the Gaussian probability distributions shown in Eq.~(\ref{eq:pn_graph}) and Eq.~(\ref{eq:ph_graph}), the above MAP estimate of Eq.~(\ref{eq:map}) can be expressed as:
\begin{equation}
\begin{split}
h_\text{MAP}(x) \coloneqq &\arg \min_h \left(-\text{log}~p_E(x-\chi h) -\text{log}~p_H(h)\right) \\
= &\arg \min_h \left(-\text{log}~e^{-(x-\chi h)^T(x-\chi h)} -\alpha~\text{log}~e^{-h^T\Lambda h}\right) \\
= &\arg \min_h ||x-\chi h||_2^2 + \alpha~h^T \Lambda h,
\end{split}
\label{eq:map_gaussian_graph}
\end{equation}
where $\alpha$ is some constant parameter proportional to the variance of the noise $\sigma_\epsilon^2$ in Eq.~(\ref{eq:pn_graph}). In a noise-free scenario where $x=\chi h$, Eq.~(\ref{eq:map_gaussian_graph}) corresponds to minimizing the following quantity:
\begin{equation}
h^T \Lambda h = (\chi^T x)^T \Lambda \chi^T x = x^T \chi \Lambda \chi^T x = x^T L x.
\label{eq:smooth}
\end{equation}
The Laplacian quadratic term in Eq.~(\ref{eq:smooth}) is the same as the one in Eq.~(\ref{eq:smoothness}). Therefore, it confirms that in a factor analysis model in Eq.~(\ref{eq:fa_graph}), a Gaussian prior in Eq.~(\ref{eq:ph_graph}) imposed on the latent variable $h$ leads to smoothness properties for the graph signal $x$. Similar observations can be made in a noisy scenario, where the main component of the signal $x$, namely, $\chi h$, is smooth on the graph. 

We are now ready to introduce the proposed learning framework. Notice that in Eq.~(\ref{eq:map_gaussian_graph}) both the representation matrix $\chi$ and the precision matrix $\Lambda$ of the Gaussian prior distribution imposed on $h$ come from the graph Laplacian $L$. They respectively represent the eigenvector and eigenvalue matrices of $L$. When the graph is unknown, we can therefore have the following joint optimization problem of $\chi$, $\Lambda$ and $h$ in order to infer the graph Laplacian:
\begin{equation}
\min_{\chi,\Lambda,h} ||x-\chi h||_2^2 + \alpha~h^T \Lambda h.
\label{eq:graph_learning_gaussian}
\end{equation}
The objective in Eq.~(\ref{eq:graph_learning_gaussian}) can be interpreted as the probability of the latent variable $h$ conditioned on the observations $x$ and the graph Laplacian L, which we consider as another variable. Using a change of variable $y=\chi h$, we thus have:
\begin{equation}
\min_{L,y} ||x-y||_2^2 + \alpha~y^T L y.
\label{eq:graph_learning_gaussian2}
\end{equation}
According to the factor analysis model in Eq.~(\ref{eq:fa_graph}), $y$ can be considered as a ``noiseless'' version of the zero-mean observation $x$. Recall that the Laplacian quadratic form $y^TLy$ in Eq.~(\ref{eq:graph_learning_gaussian2}) is usually considered as a measure of smoothness of the signal $y$ on the graph. Solving the problem of Eq.~(\ref{eq:graph_learning_gaussian2}) is thus equivalent to finding jointly the graph Laplacian $L$ and $y$, such that $y$ is close to the observation $x$, and at the same time $y$ is smooth on the learned graph. As a result, it enforces the smoothness property of the observed signals on the learned graph.

\subsection{Learning algorithm}
We propose to solve the problem in Eq.~(\ref{eq:graph_learning_gaussian2}) with the following objective function given in a matrix form:
\begin{equation}
\begin{split}
\min_{L \in \mathbb{R}^{n \times n},Y \in \mathbb{R}^{n \times p}} ~||X-Y||_F^2 + &\alpha~\mathrm{tr}(Y^T L Y) + \beta ||L||_F^2, \\
\text{s.t.} \quad & \mathrm{tr}(L)=n, \\
\quad & L_{ij} = L_{ji} \le 0, ~i \neq j, \\
\quad & L \cdot \mathbf{1}= \mathbf{0},
\end{split}
\label{eq:graph_learning_gaussian3}
\end{equation}
where $X \in \mathbb{R}^{n \times p}$ contains the $p$ input data samples $\{x_i\}_{i=1}^p$ as columns, $\alpha$ and $\beta$ are two positive regularization parameters, and $\mathbf{1}$ and $\mathbf{0}$ denote the constant one and zero vectors. In addition, $\mathrm{tr}(\cdot)$ and $||\cdot||_F$ denote the trace and Frobenius norm of a matrix, respectively. The first constraint (the trace constraint) in Eq.~(\ref{eq:graph_learning_gaussian3}) acts as a normalization factor and permits to avoid trivial solutions, and the second and third constraints guarantee that the learned $L$ is a valid Laplacian matrix that is positive semidefinite.
Furthermore, the trace constraint essentially fixes the $L^1$-norm of $L$, while the Frobenius norm is added as a penalty term in the objective function to control the distribution of the off-diagonal entries in $L$, namely, the edge weights of the learned graph. When $Y$ is fixed, the optimization problem bears similarity to the linear combination of $L^1$ and $L^2$ penalties in an elastic net regularization \cite{Zou05} in the sense that the sparsity term is imposed by the trace constraint.

The optimization problem of Eq.~(\ref{eq:graph_learning_gaussian3}) is not jointly convex in $L$ and $Y$. Therefore, we adopt an alternating minimization scheme where, at each step, we fix one variable and solve for the other variable. The solution therefore corresponds to a local minimum rather than a global minimum. Specifically, we first initialize $Y$ as the signal observations $X$. Then, at the first step, for a given $Y$, we solve the following optimization problem with respect to $L$:
\begin{equation}
\begin{split}
\min_{L} ~&\alpha~\mathrm{tr}(Y^T L Y) + \beta ||L||_F^2, \\
\text{s.t.} \quad & \mathrm{tr}(L)=n, \\
\quad & L_{ij} = L_{ji} \le 0, ~i \neq j, \\
\quad & L \cdot \mathbf{1}= \mathbf{0}.
\end{split}
\label{eq:graph_learning_gaussian_L}
\end{equation}
At the second step, $L$ is fixed and we solve the following optimization problem with respect to $Y$:
\begin{equation}
\min_{Y} ~||X-Y||_F^2 + \alpha~\mathrm{tr}(Y^T L Y).
\label{eq:graph_learning_gaussian_Y}
\end{equation}

Both problems of Eq.~(\ref{eq:graph_learning_gaussian_L}) and Eq.~(\ref{eq:graph_learning_gaussian_Y}) can be cast as convex optimization problems with unique minimizers. 
First, the problem of Eq.~(\ref{eq:graph_learning_gaussian_L}) can be written as a quadratic program (QP).
In more details, notice that the matrix $L \in \mathbb{R}^{n \times n}$ in the problem of Eq.~(\ref{eq:graph_learning_gaussian_L}) is symmetric, which means that we only need to solve for the lower triangular part of $L$, that is, the $\frac{n(n+1)}{2}$ entries on and below the main diagonal. Therefore, instead of the square matrix form, we solve for the half-vectorization of $L$ that is obtained by vectorizing the lower triangular part of $L$. We denote the half-vectorization and vectorization of $L$ as $\text{vech}(L) \in \mathbb{R}^{\frac{n(n+1)}{2}}$ and $\text{vec}(L)  \in \mathbb{R}^{n^2}$, respectively, and the former can be converted into the latter using the duplication matrix $\mathcal{M}_\text{dup}$ \cite{Abadir05}:
\begin{equation}
\mathcal{M}_\text{dup}~\text{vech}(L) = \text{vec}(L).
\label{eq:dup}
\end{equation}
Now, by using Eq.~(\ref{eq:dup}) together with the fact that:
\begin{equation}
tr(Y^T L Y) = \text{vec}(Y Y^T)^T~\text{vec}(L),
\end{equation}
and
\begin{equation}
||L||_F^2 = \text{vec}(L)^T~\text{vec}(L),
\end{equation}
we can rewrite the problem of Eq.~(\ref{eq:graph_learning_gaussian_L}) as:
\begin{equation}
\begin{split}
\arg \min_{\text{vech}(L)} ~&\alpha~\text{vec}(Y Y^T)^T~\mathcal{M}_\text{dup}~\text{vech}(L) \\
+ &\beta~\text{vech}(L)^T~\mathcal{M}_\text{dup}^T~\mathcal{M}_\text{dup}~\text{vech}(L), \\
\text{s.t.} \quad &A~\text{vech}(L) = \mathbf{0}, \\
\quad &B~\text{vech}(L) \leq \mathbf{0},
\end{split}
\label{eq:graph_learning_gaussian_L_vech}
\end{equation}
where $A$ and $B$ are the matrices that handle the equality and inequality constraints in Eq.~(\ref{eq:graph_learning_gaussian_L}). The problem of Eq.~(\ref{eq:graph_learning_gaussian_L_vech}) is a quadratic program with respect to the variable $\text{vech}(L)$ subject to linear constraints, and can be solved efficiently via interior point methods \cite{Boyd04}. As we can see, the computational complexity scales quadratically with the number of vertices in the graph. With graphs of very large number of vertices, we can instead use operator splitting methods (e.g., alternating direction method of multipliers (ADMM) \cite{Boyd11}) to find a solution. Finally, once we solve the problem of Eq.~(\ref{eq:graph_learning_gaussian_L_vech}), we convert $\text{vech}(L)$ into the square matrix form in order to solve the problem of Eq.~(\ref{eq:graph_learning_gaussian_Y}).
Second, the problem of Eq.~(\ref{eq:graph_learning_gaussian_Y}) has the following closed form solution:
\begin{equation}
Y = (I_n+\alpha L)^{-1}X.
\label{eq:chol}
\end{equation}
In practice, since the matrix $I_n+\alpha L$ is Hermitian and positive-definite, we can use the Cholesky factorization to compute $Y$ efficiently in Eq.~(\ref{eq:chol}) \cite{Boyd04}. We then alternate between these two steps of Eq.~(\ref{eq:graph_learning_gaussian_L}) and Eq.~(\ref{eq:graph_learning_gaussian_Y}) to get the final solution to the problem of Eq.~(\ref{eq:graph_learning_gaussian3}), and we generally observe convergence within a few iterations. The complete algorithm is summarized in Algorithm~\ref{alg:graph_learning_gaussian}.

\begin{algorithm}[htb]
\caption{Graph Learning for Smooth Signal Representation (\textbf{GL-SigRep})}
\begin{algorithmic}
\item [ 1:] {\bf Input:} Input signal $X$, number of iterations $\mathrm{iter}$, $\alpha$, $\beta$
\item [ 2:] {\bf Output:}  Output signal $Y$, graph Laplacian $L$
\item [ 3:] {\bf Initialization:} $Y=X$
\item [ 4:] {{\bfseries for} $t=1,2,...,\mathrm{iter}$ \bfseries{do}:}
\item [ 5:]  \quad {\bfseries Step to update Graph Laplacian $L$:} 
\item [ 6:]  \quad \quad Solve the optimization problem of Eq.~(\ref{eq:graph_learning_gaussian_L}) to update $L$.
\item [ 7:]  \quad {\bfseries Step to update $Y$:} 
\item [ 8:]  \quad \quad Solve the optimization problem of Eq.~(\ref{eq:graph_learning_gaussian_Y}) to update $Y$.
\item [ 9:] {\bf end for}  
\item [ 10:]  $L=L^{\mathrm{iter}}, Y=Y^{\mathrm{iter}}$.
\end{algorithmic}
\label{alg:graph_learning_gaussian}
\end{algorithm}

\section{Experiments}
\label{sec:experiment}
In this section, we evaluate the performance of the proposed graph learning algorithm. We first describe the general experimental setting, and then we present experimental results on both synthetic and real world data.

\subsection{Experimental settings}
We test the performance of our framework by comparing the graphs learned from sets of synthetic or real world observations to the groundtruth graphs. We provide both visual and quantitative comparisons, where we compare the existence of edges in the learned graph to the ones of the groundtruth graph. We use four evaluation criteria commonly used in information retrieval, namely, \textit{Precision}, \textit{Recall}, \textit{F-measure} and \textit{Normalized Mutual Information (NMI)} \cite{Manning08}, to test the performance of our algorithm. For computing the \textit{NMI}, we first compute a 2-cluster partition of all the vertex pairs using the learned graph, based on whether or not there exists an edge between the two vertices. We then compare this partition with the 2-class partition obtained in the same way using the groundtruth graph. 

In our experiments, we solve the optimization problem of Eq.~(\ref{eq:graph_learning_gaussian_L}) using the convex optimization package CVX \cite{Grant13,Grant08}. Our algorithm $\textbf{GL-SigRep}$ stops when the maximum number of iterations is reached or the absolute change in the objective is smaller than $10^{-4}$. In most of the cases, we observed that the algorithm converges within a few iterations. The experiments are carried out on different sets of parameters, namely, for different values of $\alpha$ and $\beta$ in Eq.~(\ref{eq:graph_learning_gaussian3}). Finally, we prune insignificant edges that have a weight smaller than $10^{-4}$ in the learned graph.

We compare the proposed graph learning framework to a state-of-the-art machine learning approach for estimating graph structure, which is closely related to the idea of sparse inverse covariance estimation for GMRF. Specifically, Lake et al. propose in \cite{Lake10} to solve the following optimization problem:
\begin{equation}
\begin{split}
\min_{L_\text{pre} \succ 0, W, \sigma^2} ~&\mathrm{tr}(L_\text{pre} \frac{1}{p} X X^T) - \text{log}|L_\text{pre}| + \frac{\lambda}{p} || W ||_1, \\
\text{s.t.} \quad & L_\text{pre} = \text{diag}(\sum_j w_{ij}) - W + I / \sigma^2, \\
\quad & w_{ii} = 0, ~i = 1, \dots, n, \\
\quad & w_{ij} \geq 0, ~i = 1, \dots, n;~j = 1, \dots, n, \\
\quad & \sigma^2 > 0,
\end{split}
\label{eq:logdet}
\end{equation}
where $W$ is the adjacency matrix of the target graph with edge weight $w_{ij}$ between vertices $i$ and $j$, $L_\text{pre}$ is a precision matrix for the multivariate Gaussian distribution that generates the observations $X$, $\lambda$ is a regularization parameter, and $| \cdot |$ and $|| \cdot ||_1$ denote the determinant and the $L^1$-norm, respectively. It can be seen that the objective function in Eq.~(\ref{eq:logdet}) is very similar to the $L^1$ regularized log-determinant program proposed in \cite{Banerjee08,Friedman08}; however, by adding the constraints in Eq.~(\ref{eq:logdet}), Lake et al. essentially defines the precision matrix as a regularized graph Laplacian with positive diagonal loading, where the parameter $\sigma^2$ is interpreted as the a priori feature variance \cite{Zhu03}. As a consequence, by solving the problem of Eq.~(\ref{eq:logdet}), they are able to obtain a valid Laplacian matrix, which is not the case for the methods in \cite{Banerjee08,Friedman08}. Therefore, we choose this method instead of the ones in \cite{Banerjee08,Friedman08} for a more fair comparison. In our experiments we solve the problem of Eq.~(\ref{eq:logdet}) using CVX, similarly to what is indicated in \cite{Lake10}. We denote this algorithm as \textbf{GL-LogDet}. We test \textbf{GL-LogDet} based on different choices of the parameter $\lambda$ in Eq.~(\ref{eq:logdet}). After obtaining the Laplacian matrix, we prune insignificant connections that have a weight smaller than $10^{-4}$, similarly to the post-processing done with $\textbf{GL-SigRep}$.

\subsection{Results on synthetic data}
We first carry out experiments on three different synthetic graphs of 20 vertices, namely, a graph whose edges are determined based on Euclidean distances between vertices, and two graphs that follow the Erd\H{o}s-R\'{e}nyi model \cite{Erdos60} and the Barab\'{a}si-Albert model \cite{Barabasi99}, respectively. For the first graph, we generate the coordinates of the vertices uniformly at random in the unit square, and compute the edge weights with a Gaussian radial basis function (RBF), i.e., $\text{exp}\left(-d(i,j)^2/2 \sigma^2\right)$ where $d(i,j)$ is the distance between vertices and $\sigma=0.5$ is a kernel width parameter. We then remove all the edges whose weights are smaller than 0.75. Next, we use the Erd\H{o}s-R\'{e}nyi (ER) model with edge probability 0.2 to generate a random graph, that is, each possible edge is included in the graph with probability 0.2 independently of other edges. Finally, we use the Barab\'{a}si-Albert (BA) model to generate a scale-free random graph. Specifically, the BA graph in our experiments is generated by adding one new vertex to the graph at each time, connecting to one existing vertex in the graph. The probability of the new vertex to be linked to a given existing vertex in the graph is proportional to the ratio of the degree of that existing vertex to the sum of degrees of all the existing vertices. The ER and BA graphs are important random graph models studied in network science. Specifically, the former has power-law (or scale-free) degree distributions similarly to many networks observed in real world, while the latter does not. The ER and BA graphs in our experiments have unitary edge weights.

Given a groundtruth synthetic graph, we compute the graph Laplacian $L$ and normalize the trace according to Eq.~(\ref{eq:graph_learning_gaussian3}). Then, for each graph, we generate 100 signals $X=\{x_i\}_{i=1}^{100}$ that follow the distribution shown in Eq.~(\ref{eq:px_graph}) with $u_x = 0$ and $\sigma_\epsilon = 0.5$. We then apply \textbf{GL-SigRep} and \textbf{GL-LogDet} to learn the graph Laplacian matrices, given only the signals $X$. 
Finally, we also compute as a baseline the sample correlation matrix $S$ where $S_{ij}$ represents the Pearson product-moment correlation coefficient between signal observations at vertices $i$ and $j$.

\begin{figure*}[!tbp]
      \centering
            \subfigure[Gaussian: Groundtruth]		{ \includegraphics[width=3.2cm]{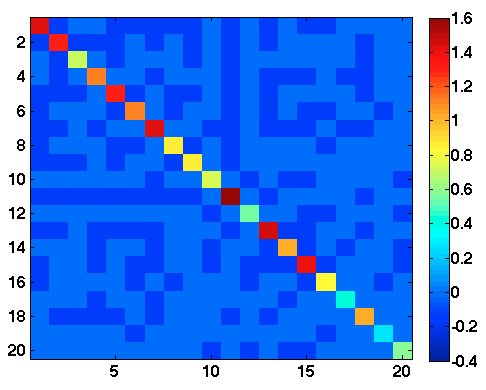}   		\label{laplacian_gt_rbf}}
            \subfigure[Gaussian: \textbf{GL-SigRep}] 	{ \includegraphics[width=3.2cm]{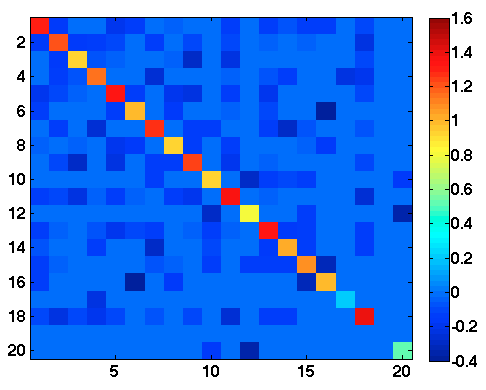}		\label{laplacian_sigrep_rbf}}
            \subfigure[Gaussian: \textbf{GL-LogDet}] 	{ \includegraphics[width=3.2cm]{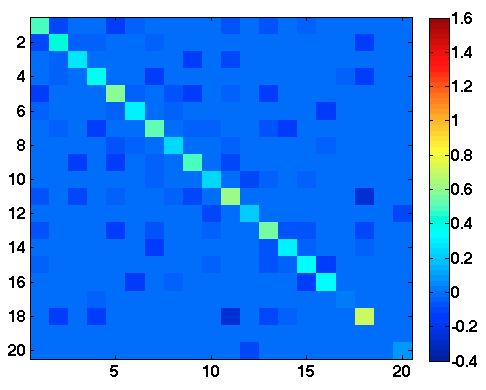}		\label{laplacian_logdet_rbf}}
            \subfigure[Gaussian: Sample correlation] 	{ \includegraphics[width=3.2cm]{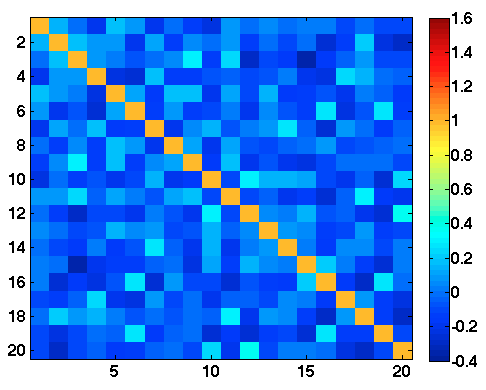} 	\label{sample_corr_rbf}}
            \subfigure[Gaussian: Sample correlation (thresholded)] 	{ \includegraphics[width=3.2cm]{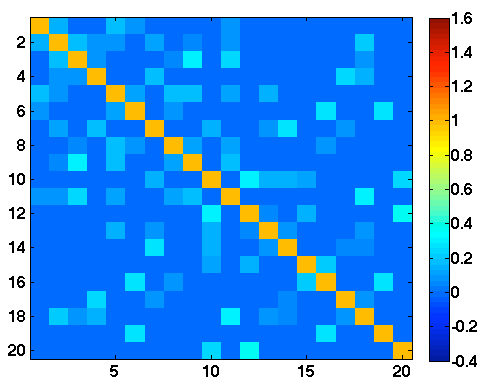} 	\label{sample_corr_rbf}}
            \subfigure[ER: Groundtruth]			{ \includegraphics[width=3.2cm]{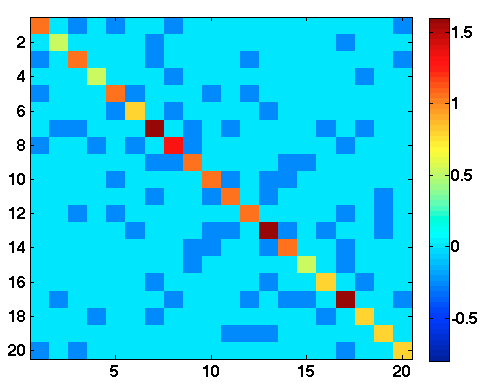} 			\label{laplacian_gt_er}}
            \subfigure[ER: \textbf{GL-SigRep}]		{ \includegraphics[width=3.2cm]{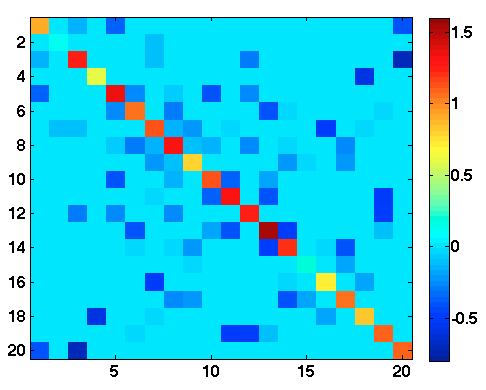}   	\label{laplacian_sigrep_er}}
            \subfigure[ER: \textbf{GL-LogDet}]		{ \includegraphics[width=3.2cm]{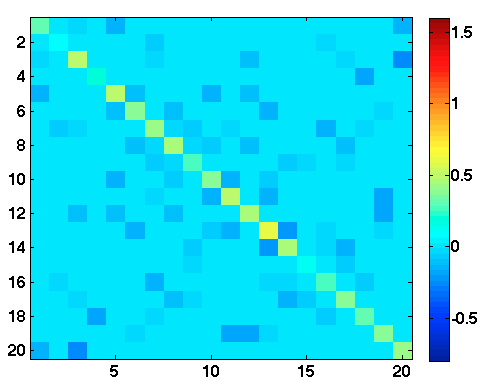}		\label{laplacian_logdet_er}}
            \subfigure[ER: Sample correlation]		{ \includegraphics[width=3.2cm]{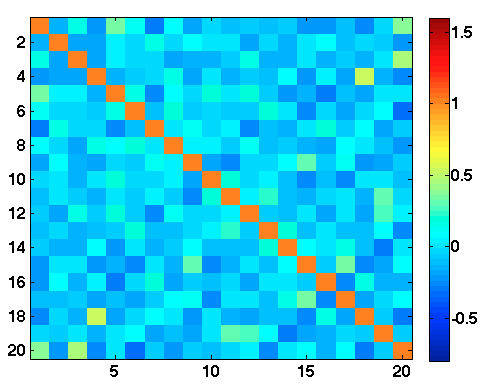}		\label{sample_corr_er}}
            \subfigure[ER: Sample correlation (thresholded)]		{ \includegraphics[width=3.2cm]{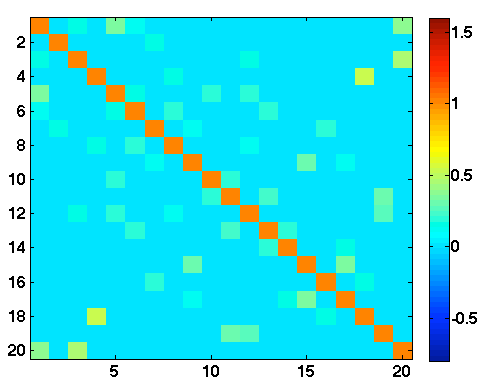}		\label{sample_corr_er}}
            \subfigure[BA: Groundtruth]			{ \includegraphics[width=3.2cm]{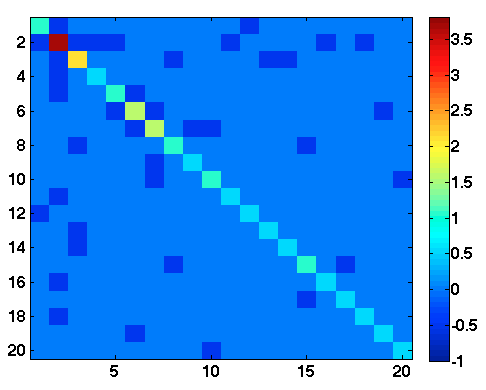}			\label{laplacian_gt_ba}}
            \subfigure[BA: \textbf{GL-SigRep}]		{ \includegraphics[width=3.2cm]{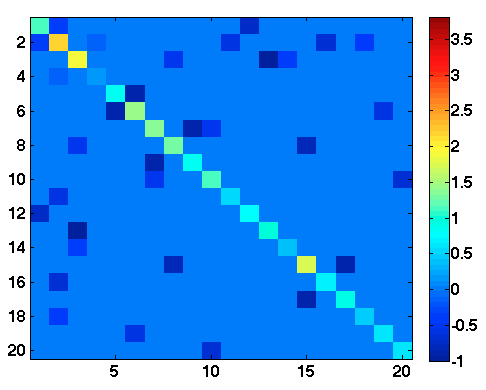} 		\label{laplacian_sigrep_ba}}
            \subfigure[BA: \textbf{GL-LogDet}]		{ \includegraphics[width=3.2cm]{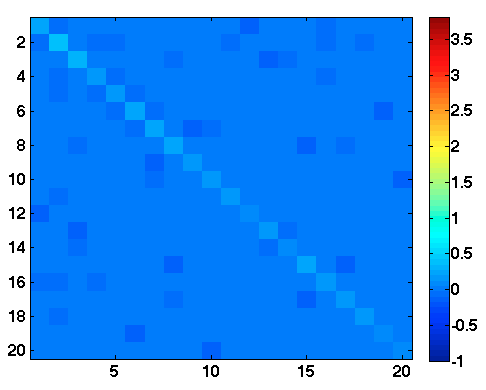}   	\label{laplacian_logdet_ba}}
            \subfigure[BA: Sample correlation]		{ \includegraphics[width=3.2cm]{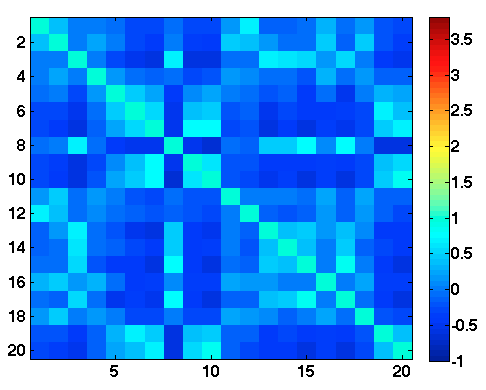}		\label{sample_corr_ba}}
            \subfigure[BA: Sample correlation (thresholded)]		{ \includegraphics[width=3.2cm]{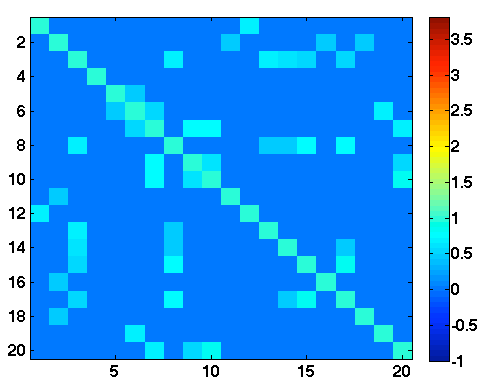}		\label{sample_corr_ba}}
        \vspace{-0.2cm}
        \caption{The learned graph Laplacian matrices as well as sample correlation matrices (original and thresholded). The columns from the left to the right are the groundtruth Laplacians, the Laplacians learned by \textbf{GL-SigRep}, the Laplacians learned by \textbf{GL-LogDet}, the sample correlation matrices, and the thresholded sample correlation matrices. The rows from the top to the bottom are for the Gaussian RBF graph, the ER graph, and the BA graph.}
        \label{fig:syn_visual}
\end{figure*}

We first provide visual comparisons in Fig.~\ref{fig:syn_visual}, where we show from the left to the right columns the Laplacian matrices of the groundtruth graph, the graph Laplacians learned by \textbf{GL-SigRep}, the graph Laplacians learned by \textbf{GL-LogDet}, 
the sample correlation matrices, and the thresholded sample correlation matrices,
for one random instance of each of the three graph models\footnote{These results are obtained based on the parameters, namely, $\alpha$ and $\beta$ in \textbf{GL-SigRep}, $\lambda$ in \textbf{GL-LogDet}, and threshold for the thresholded sample correlation matrix, that lead to the best \textit{F-measure} scores (see quantitative results in the next paragraph). More discussion about the choices of these parameters are presented later.}. 
In each row, the values in the Laplacian and correlation matrices are scaled to be within the same range. 
First, we see that the sample correlation matrix simply computes pairwise correlations and therefore may lead to undesirable noisy correlation values as shown in Fig.~\ref{fig:syn_visual}(d), (i), (n). The thresholded sample correlation matrices show that the positions of the large positive entries in the sample correlation matrices generally correspond to the positions of the edges in the groundtruth graph; however, the intensities of these ``edges'' are less consistent with the groundtruth than those in the Laplacian matrices learned by our method, as we see later. One possible reason for this is that the sample correlation matrix computes pairwise correlations between (possibly noisy) observations; in comparison, with a global smoothness constraint in our model, we rather learn a graph topology such that the signals are regularized to be globally smooth.
Second, we see that for all three types of graphs, the graph Laplacian matrices learned by \textbf{GL-SigRep} are visually more consistent with the groundtruth data than the ones learned by \textbf{GL-LogDet}, especially in the case of a BA graph.

\begin{table}[t]
\caption{Graph learning performance for \textbf{GL-SigRep}, \textbf{GL-LogDet} and sample correlation matrices.}
\centering
\subtable[The Gaussian RBF graph.]{
\begin{tabular}{c c c c c}
\hline\hline
Algorithm& F-measure & Precision & Recall  & NMI \\ [0.5ex] 
\hline
\textbf{GL-SigRep}& 0.8480 & 0.8050 & 0.9007 & 0.5146 \\
\textbf{GL-LogDet}& 0.8159 & 0.8100 & 0.8345 & 0.4632 \\
\textbf{Samp. Corr.}& 0.8193 & 0.8174 & 0.8361 & 0.4659 \\[1ex]
\hline
\end{tabular}
} \\
\subtable[The ER graph.]{
\begin{tabular}{c c c c c}
\hline\hline
Algorithm& F-measure & Precision & Recall  & NMI \\ [0.5ex] 
\hline
\textbf{GL-SigRep}& 0.7236 & 0.6391 & 0.8447 & 0.3695 \\
\textbf{GL-LogDet}& 0.7246 & 0.6273 & 0.8737 & 0.3903 \\
\textbf{Samp. Corr.}& 0.7019 & 0.7846 & 0.6536 & 0.3725 \\[1ex]
\hline
\end{tabular}
} \\
\subtable[The BA graph.]{
\begin{tabular}{c c c c c}
\hline\hline
Algorithm& F-measure & Precision & Recall  & NMI \\ [0.5ex] 
\hline
\textbf{GL-SigRep}& 0.9342 & 0.9514 & 0.9211 & 0.8366 \\
\textbf{GL-LogDet}& 0.8387 & 0.8113 & 0.9158 & 0.6783 \\
\textbf{Samp. Corr.}& 0.6819 & 0.8149 & 0.6105 & 0.4380 \\[1ex]
\hline
\end{tabular}
}
\label{tab:syn_quan}
\vspace*{-0.5cm}
\end{table}

\begin{figure*}
      \centering
            \subfigure[]	{ \includegraphics[width=5cm]{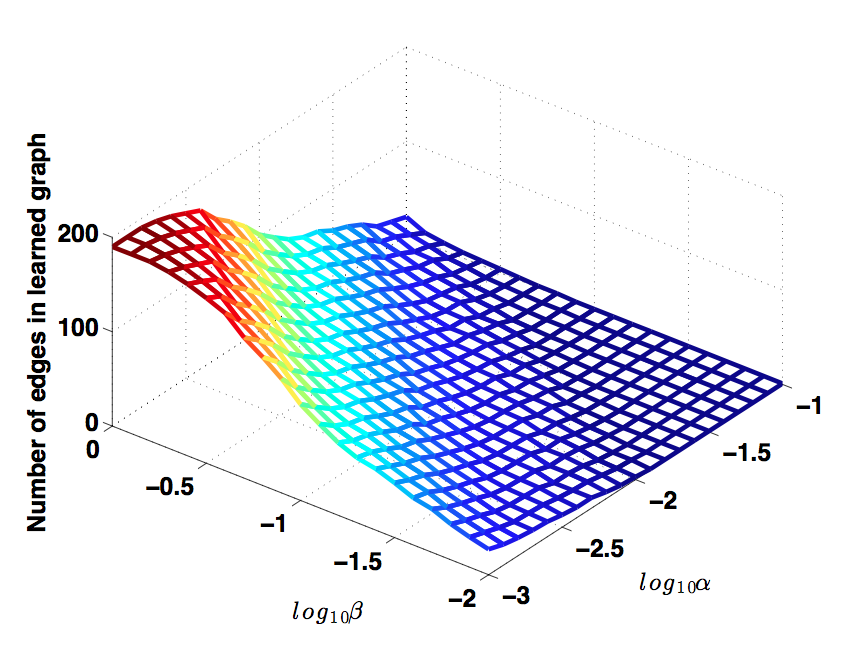}			\label{num_edges_trend}}
            \subfigure[]	{ \includegraphics[width=5cm]{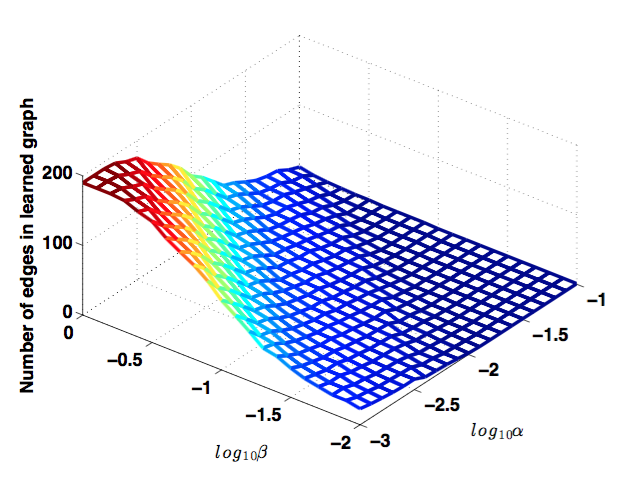}		\label{num_edges_trend_er}}
            \subfigure[]	{ \includegraphics[width=5cm]{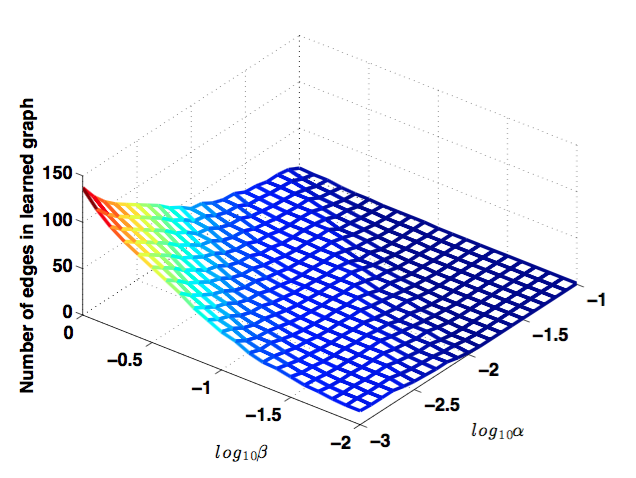}		\label{num_edges_trend_ba}}
            \subfigure[]	{ \includegraphics[width=5cm]{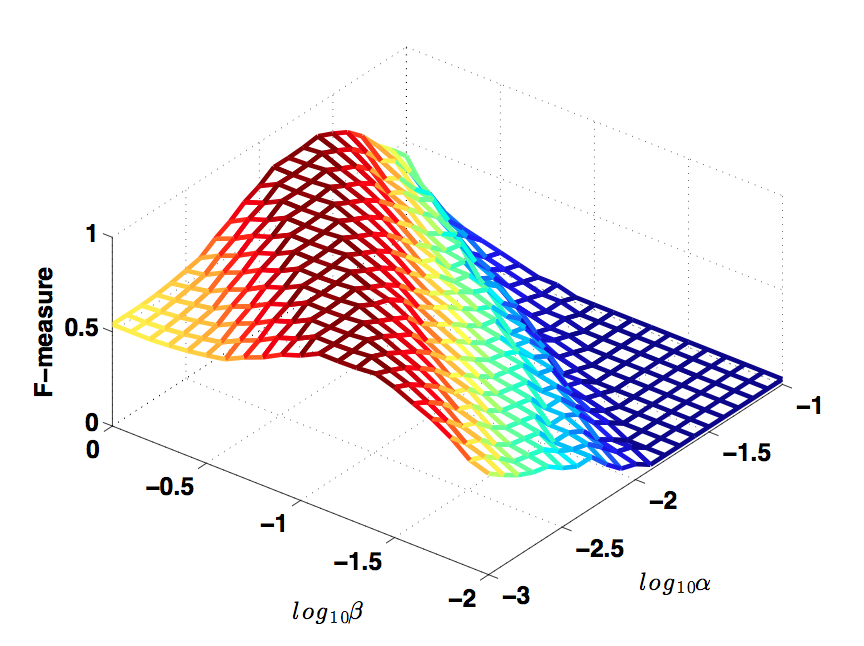}					\label{f_trend}}
            \subfigure[]	{ \includegraphics[width=5cm]{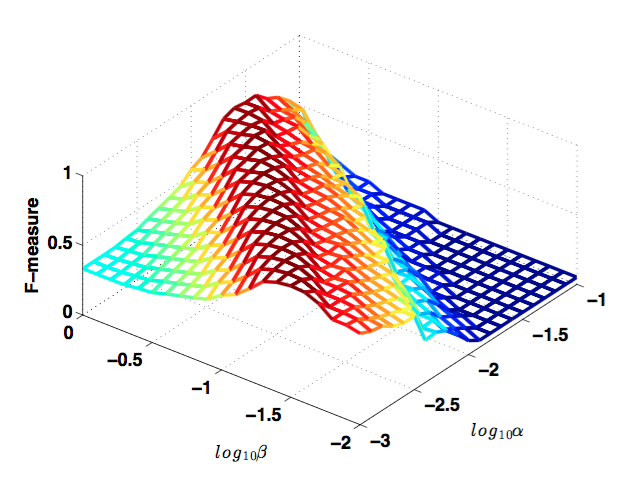}				\label{f_trend_er}}
            \subfigure[]	{ \includegraphics[width=5cm]{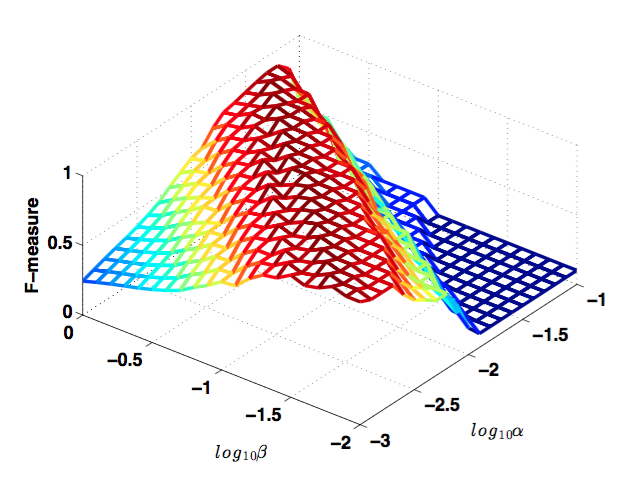}				\label{f_trend_ba}}
        \vspace{-0.2cm}
        \caption{(a-c) The number of edges in the learned graph, and (d-f) the \textit{F-measure} score, under 441 different combinations of the parameters $\alpha$ and $\beta$, for a random instance of the Gaussian RBF graph (left column), ER graph (middle column), and BA graph (right column).}
        \label{fig:syn_sigrep_trend}
\end{figure*}

Next, we quantitatively evaluate the performance of \textbf{GL-SigRep}, \textbf{GL-LogDet}, and sample correlations, in terms of recovering the position of the edges in the groundtruth graph. 
For sample correlation matrices, we consider off-diagonal correlations with values greater than a chosen threshold (see Footnote 4 for the exact values) as the ``recovered edges''.
The evaluation metrics we choose are the \textit{F-measure}, \textit{Precision}, \textit{Recall} and \textit{Normalized Mutual Information (NMI)} \cite{Manning08}. Specifically, the \textit{Precision} evaluates the percentage of correct edges in the learned graph, that is, the edges that are present in the groundtruth graph. The \textit{Recall} evaluates the percentage of the edges in the groundtruth graph that are present in the learned graph. The \textit{F-measure} thus takes into account both \textit{Precision} and \textit{Recall} to measure the overall accuracy of the obtained edge set. Finally, the \textit{NMI} measures the mutual dependence between the obtained edge set and that of the groundtruth graph from an information theoretic viewpoint.
Testing on different parameter values in the three methods\footnote{We choose parameters through a grid search. The optimal values for $\alpha$, $\beta$, $\lambda$, and threshold for the sample correlation matrix, are 0.012, 0.79, 19.95 and 0.06, respectively, for the Gaussian RBF graph; 0.0032, 0.10, 12.59 and 0.10, respectively, for the ER graph; and 0.0025, 0.050, 125.89 and 0.46, respectively, for the BA graph.}, we show in Table~\ref{tab:syn_quan} the best \textit{F-measure}, \textit{Precision}, \textit{Recall} and \textit{NMI} scores achieved by the three methods averaged over ten random instances of the three graphs with the associated signals $X$. Our algorithm \textbf{GL-SigRep} provides competitive or superior performance compared to the other two in terms of all the evaluation criteria. Especially, for the Gaussian RBF and BA graphs, the high average \textit{F-measure} scores achieved by \textbf{GL-SigRep} suggest that the learned graphs have topologies that are very similar to the groundtruth ones. 
The advantage of \textbf{GL-SigRep} for the non-manifold structured ER graphs is less obvious, possibly due to the fact that the edges are generated independently in this random graph model. This may lead to a weaker global relationship in the data and therefore affects the efficiency of a global relationship model like ours.
Finally, we have also computed the mean squared error (MSE) between the groundtruth and the learned Laplacian matrices in order to study the accuracy of the edge weights. For the sample correlation method, we construct an adjacency matrix where the edges correspond to the correlations above the chosen threshold (the ones in Footnote 4), and then compute accordingly the Laplacian matrix. The MSE is 1.4345 for \textbf{GL-SigRep} compared to 2.8443 for \textbf{GL-LogDet} and 2.0508 for sample correlation in case of the Gaussian RBF graph, 2.1737 compared to 3.4350 and 3.0117 in case of the ER graph, and 3.0412 compared to 6.1491 and 5.2262 in case of the BA graph. This shows that, in addition to edge positions, the edge weights learned by \textbf{GL-SigRep} are more consistent with the grountruth than those of the other methods.


To better understand the behavior of \textbf{GL-SigRep} under different sets of parameters, we plot in Fig.~\ref{fig:syn_sigrep_trend}(a) the numbers of edges in the learned graph, and in Fig.~\ref{fig:syn_sigrep_trend}(d) the \textit{F-measure} scores, under $21\times21=441$ different combinations of the parameters $\alpha$ and $\beta$ in Eq.~(\ref{eq:graph_learning_gaussian3}), for a random instance of the Gaussian RBF graph. First, we see that the number of edges in the learned graph decreases as $\beta$ decreases and $\alpha$ increases. The intuitions behind this behavior are as follows. When $\beta$ increases, the Frobenius norm of $L$ in the objective function in Eq.~(\ref{eq:graph_learning_gaussian3}) tends to be small. Given a fixed $L^1$-norm of $L$ (which is enforced by the trace constraint), this leads to a more uniform distribution of the off-diagonal entries (with similar but smaller values), so that the number of edges tends to increase. Decreasing $\beta$ leads to the opposite effect. On the other hand, when $\alpha$ increases, the trace of the quadratic term tends to be small. 
The algorithm thus favors a smaller number of non-zero entries in $L$, and the number of edges decreases. Therefore, both parameters $\alpha$ and $\beta$ implicitly affect the sparsity of the learned graph Laplacian. It is interesting to notice in Fig.~\ref{fig:syn_sigrep_trend}(a) and Fig.~\ref{fig:syn_sigrep_trend}(b) that, both the number of edges and the \textit{F-measure} scores are similar for the values of $\alpha$ and $\beta$ that lead to the same ratio $\frac{\beta}{\alpha}$. The same patterns are also observed for the ER graph model (Fig.~\ref{fig:syn_sigrep_trend}(b)(e)) and the BA graph model (Fig.~\ref{fig:syn_sigrep_trend}(c)(f)).
This suggests that, when the initial observations are relatively smooth, the trace of the quadratic term and the Frobenius norm are the dominating factors in the optimization of Eq.~(\ref{eq:graph_learning_gaussian3}), rather than the data fidelity term. This implies that, in practice, we may only need to search for an appropriate ratio $\frac{\beta}{\alpha}$ to maximize the learning performance of the algorithm.

\begin{figure*}[t]
      \centering
            \subfigure[]	{ \includegraphics[width=6cm]{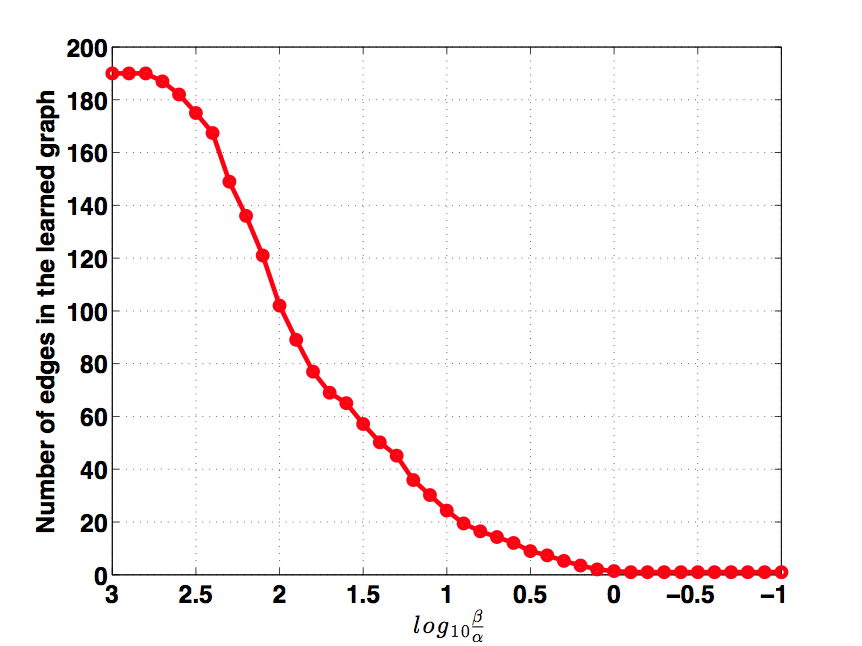}		\label{edges_gaussian_rbf}}
            \subfigure[]	{ \includegraphics[width=6cm]{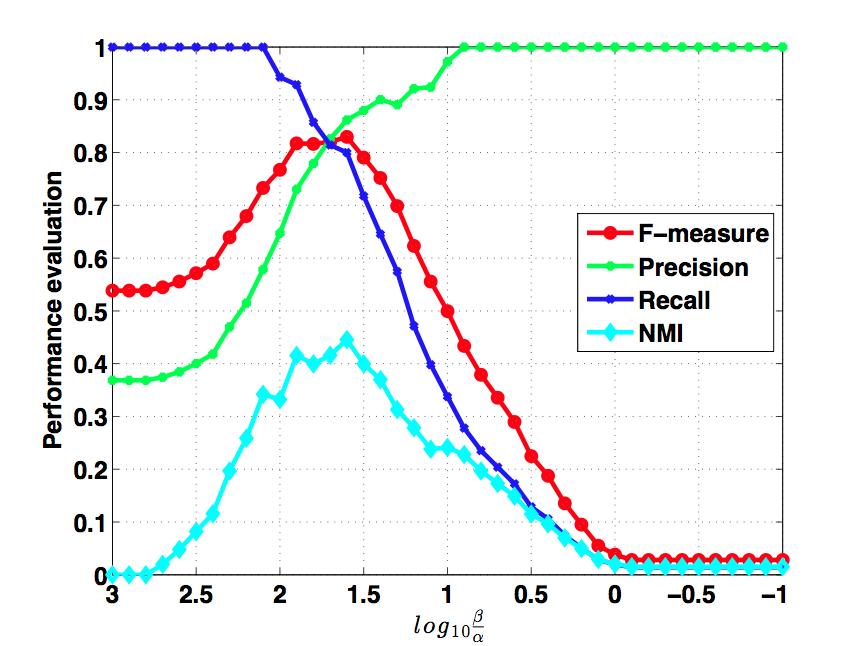}		\label{perf_gaussian_rbf}}
        \vspace{-0.2cm}
        \caption{(a) Number of edges in the graphs learned by \textbf{GL-SigRep}, and (b) performance of \textbf{GL-SigRep}, for different ratios $\frac{\beta}{\alpha}$ for a random instance of the Gaussian RBF graph.}
        \label{fig:syn_edges_perf_rbf}
\end{figure*}

\begin{figure*}[t]
      \centering
            \subfigure[]	{ \includegraphics[width=6cm]{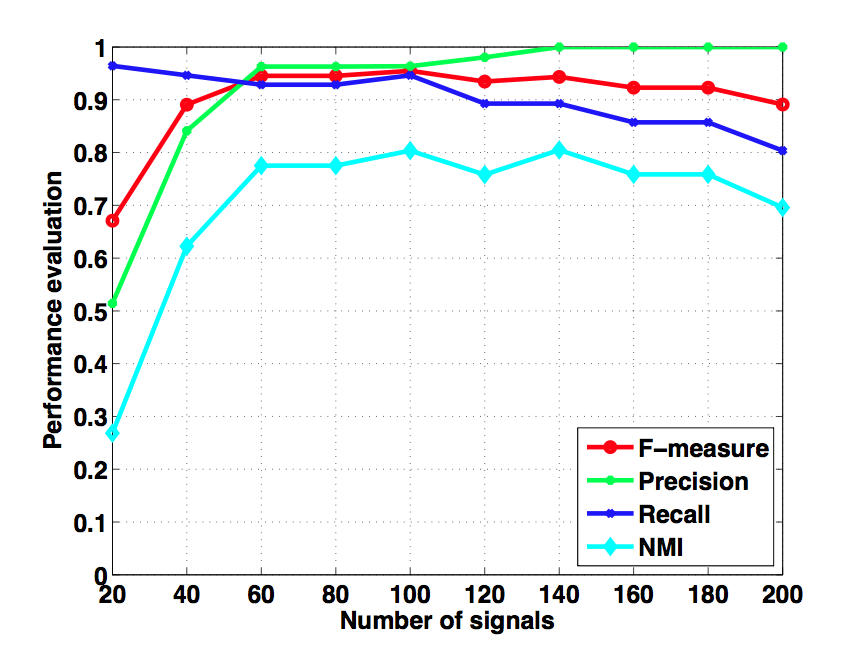}		\label{num_signal_gaussian}}
            \subfigure[]	{ \includegraphics[width=6cm]{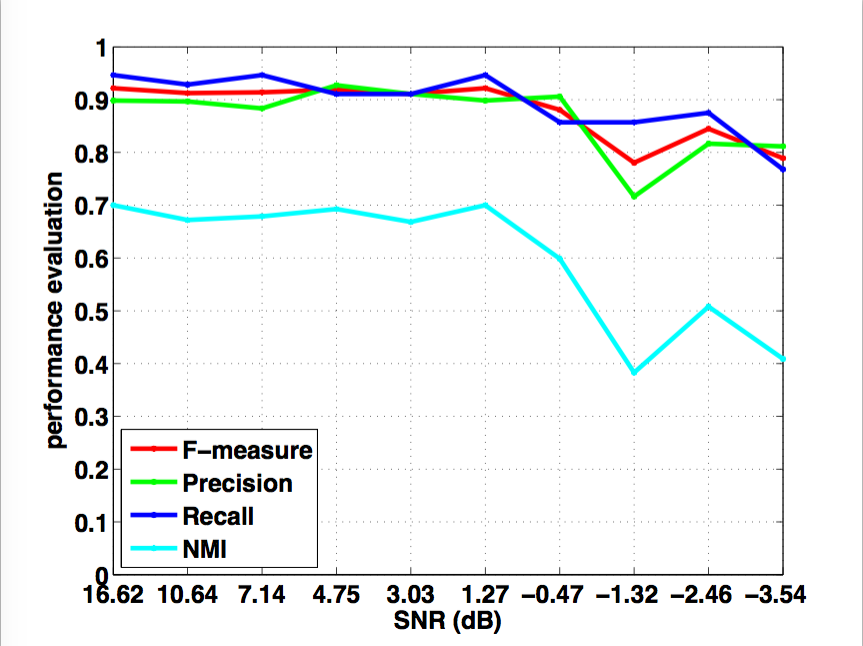}		\label{noise_level_gaussian_db}}
        \vspace{-0.2cm}
        \caption{Performance of \textbf{GL-SigRep} for (a) different numbers of signals $p$ and (b) different SNR (dB), for two random instances of the Gaussian RBF graph, respectively.}
        \label{fig:syn_num_signal_noise_level}
\end{figure*}

Next, we show in Fig.~\ref{fig:syn_edges_perf_rbf}(a) and Fig.~\ref{fig:syn_edges_perf_rbf}(b) the number of edges in the graphs learned by \textbf{GL-SigRep}, and the learning performance evaluated based on the four criteria, respectively, for the same Gaussian RBF graph as before but under different ratios of $\beta$ to $\alpha$. As expected, the number of edges decreases as the ratio of $\beta$ to $\alpha$ decreases. Looking at Fig.~\ref{fig:syn_edges_perf_rbf}(a) and Fig.~\ref{fig:syn_edges_perf_rbf}(b) together, we see that, as the number of edges approaches the number of edges in the groundtruth graph (in this case, 70 edges), the \textit{Recall} stays high and the \textit{Precision} increases rapidly, which makes the \textit{F-measure} increase. When the number of edges in the learned graph is close to the one in the groundtruth graph, the curves for the \textit{Precision} and the \textit{Recall} intersect and the \textit{F-measure} reaches its peak. After that, although the \textit{Precision} keeps increasing towards 1, the \textit{Recall} drops rapidly as fewer and fewer edges are detected, leading to a decreasing trend in the \textit{F-measure}. A similar trend can be observed in the curve for the \textit{NMI} score. The patterns in Fig.~\ref{fig:syn_edges_perf_rbf} are also observed for the ER and BA graph models. These results show that \textbf{GL-SigRep} is able to learn a graph that is very close to the groundtruth graph when the number of edges matches the number of edges in the groundtruth graph.

Finally, we investigate the influence of the number of signals available for learning, and the level of noise present in the data, for two random instances of the Gaussian RBF graph, respectively. In Fig.~\ref{fig:syn_num_signal_noise_level}(a), we show the performance of \textbf{GL-SigRep} for different numbers of signals. As we can see, the performance initially increases as more signals are available to learn the graph Laplacian, but remains quite stable when more than 60 signals are available. In Fig.~\ref{fig:syn_num_signal_noise_level}(b), we show the performance of our algorithm for different values of the signal-to-noise-ratio (SNR) given the Gaussian noise $\sigma_\epsilon$. We see that the performance of \textbf{GL-SigRep} again remains stable until the SNR becomes very low.

\subsection{Learning meteorological graph from temperature data}
\begin{figure*}
      \centering
            \subfigure[]	{ \includegraphics[width=4.2cm]{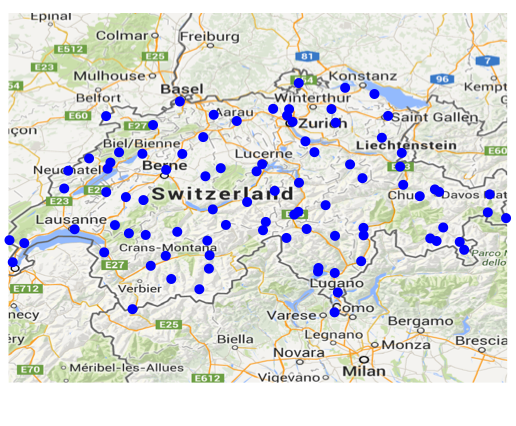}   			\label{station}}
            \subfigure[]	{ \includegraphics[width=4.2cm]{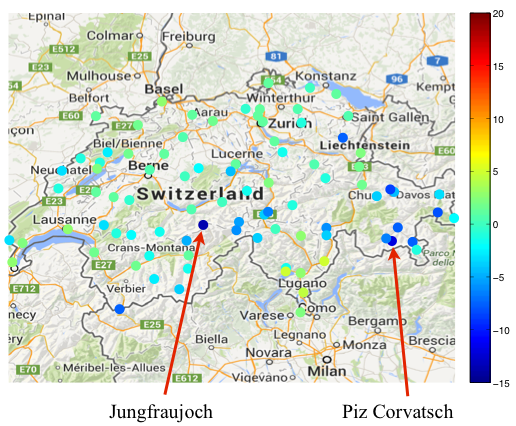}   		\label{temp_feb}}
            \subfigure[]	{ \includegraphics[width=4.2cm]{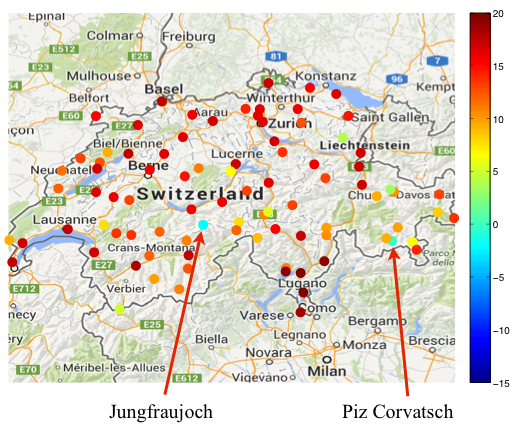}			\label{temp_jun}}
            \subfigure[]	{ \includegraphics[width=4.2cm]{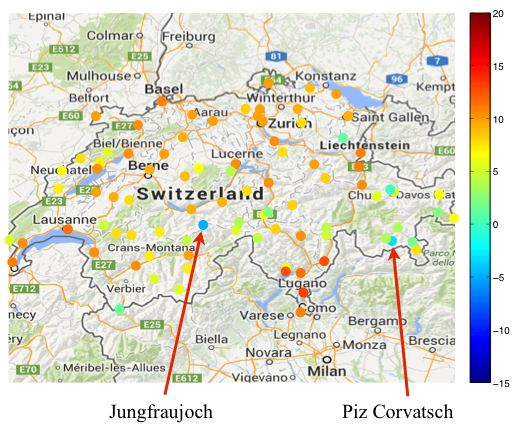}			\label{temp_oct}}
        \vspace{-0.2cm}
        \caption{(a) The locations of 89 measuring stations in Switzerland. (b)-(d) Average monthly temperature in February, June and October, respectively, in Switzerland during the period from 1981 to 2010. The color code in (b)-(d) represents the temperature in Celsius scale.}
        \label{fig:real_temp}
\end{figure*}

We now test the proposed graph learning framework on real world data. We first consider monthly temperature data collected at 89 measuring stations in Switzerland (shown in Fig.~\ref{fig:real_temp}(a)) during the period between 1981 and 2010 \cite{meteoswiss3}. Specifically, for each station, we compute the average temperature for each month over the 30 years. This leads to 12 signals that measure the average temperatures at the 89 measuring stations in the 12 months. By applying the proposed graph learning algorithm, we would like to infer a graph where stations with similar temperature evolutions across the year are connected. In other words, we aim at learning a graph on which the observed temperature signals are smooth. In this case, the natural choice of a geographical graph based on physical distances between the stations does not seem appropriate for representing the similarity of temperature values between these stations. Indeed, Fig.~\ref{fig:real_temp}(b)-(d) show the average temperatures in Switzerland in February, June and October, and we can see that the evolution of temperatures at most of the stations follows very similar trends, which implies that the stations' values are highly correlated, regardless of the geographical distances between them. On the other hand, it turns out that altitude is a more reliable source of information to determine temperature evolutions. For instance, as we observed from the data, the temperatures at two stations, Jungfraujoch and Piz Corvatsch, follow similar trends that are clearly different from other stations, possibly due to their similar altitudes (both are more than 3000 metres above sea level). Therefore, the goal of our learning experiments is then to recover a graph that reflects the altitude relationships between the stations given the observed temperature signals.

We apply the two graph learning algorithms \textbf{GL-SigRep} and \textbf{GL-LogDet} on the temperature signals to learn two graph Laplacian matrices. For quantitative evaluation, we build a groundtruth graph that reflects the similarity between stations in terms of their altitudes. More specifically, we connect two stations with a unitary weight if and only if their altitude difference is smaller than 300 metres. We then compare the groundtruth and learned graph Laplacians using the same evaluation metrics as in the synthetic experiments.

\begin{figure*}
      \centering
            \subfigure[Groundtruth]			{ \includegraphics[width=4.5cm]{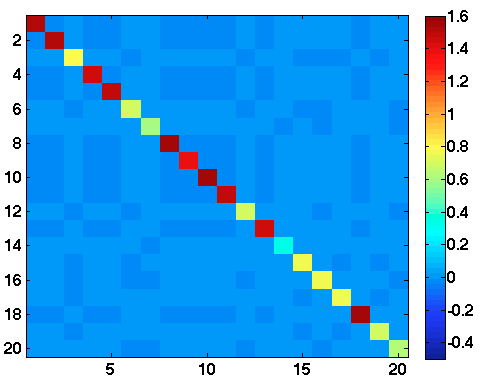}   			\label{laplacian_gt}}
            \subfigure[\textbf{GL-SigRep}]	{ \includegraphics[width=4.5cm]{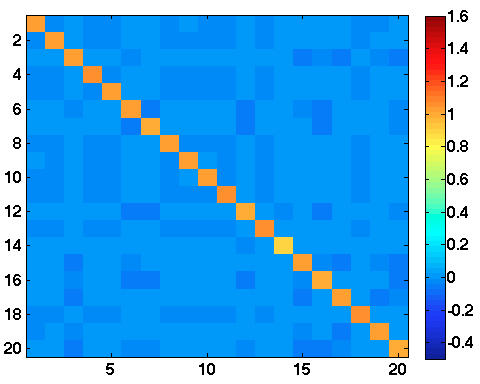}		\label{laplacian_sigrep}}
            \subfigure[\textbf{GL-LogDet}]	{ \includegraphics[width=4.5cm]{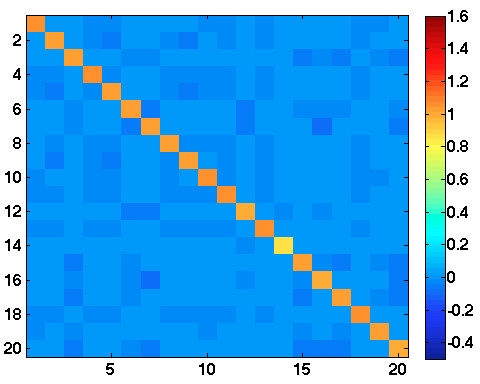}		\label{laplacian_logdet}}
        \vspace{-0.2cm}
        \caption{Visual comparisons between (a) the groundtruth graph Laplacian, (b) the graph Laplacian learned by \textbf{GL-SigRep}, and (c) the graph Laplacian learned by \textbf{GL-LogDet}. The results for \textbf{GL-SigRep} and \textbf{GL-LogDet} are obtained based on the parameters $\alpha$, $\beta$ and $\lambda$ that lead to the best \textit{F-measure} scores. The values in the graph Laplacian learned by \textbf{GL-LogDet} is scaled so that it has the same trace as the other two matrices.}
        \label{fig:real_visual}
\end{figure*}

We first show visual comparisons between the Laplacian of the groundtruth altitude-based graph and the graph Laplacians learned by \textbf{GL-SigRep} and \textbf{GL-LogDet}. For a more clear visualization, we focus on the top left part of the three matrices and plot them in Fig.~\ref{fig:real_visual}. The comparisons between \textbf{GL-SigRep} and \textbf{GL-LogDet} show that the edges in the graph learned by our algorithm are again more consistent with the groundtruth data. This is confirmed by the results shown in Table~\ref{tab:real_quan} where the best \textit{F-measure}, \textit{Precision}, \textit{Recall} and \textit{NMI} scores achieved by the two algorithms are presented. Overall, both visual and quantitative comparisons show that \textbf{GL-SigRep} achieves better performance than \textbf{GL-LogDet} in inferring the graph topology from the signal observations in this example.

\begin{table}[t]
\caption{Performance for \textbf{GL-SigRep} and \textbf{GL-LogDet} in learning the meteorological graph.}
\centering
\begin{tabular}{c c c c c}
\hline\hline
Algorithm& F-measure & Precision & Recall  & NMI \\ [0.5ex] 
\hline
\textbf{GL-SigRep}& 0.8374 & 0.7909 & 0.8896 & 0.4910 \\
\textbf{GL-LogDet}& 0.8052 & 0.7718 & 0.8417 & 0.4249 \\[1ex]
\hline
\end{tabular}
\label{tab:real_quan}
\vspace*{-0.5cm}
\end{table}

\begin{figure*}[t]
      \centering
            \subfigure[]	{ \includegraphics[width=5.4cm]{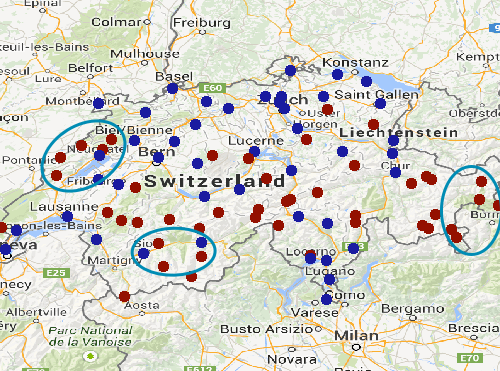}		\label{altitude_cluster_new_labeled}}
            \subfigure[]	{ \includegraphics[width=5.4cm]{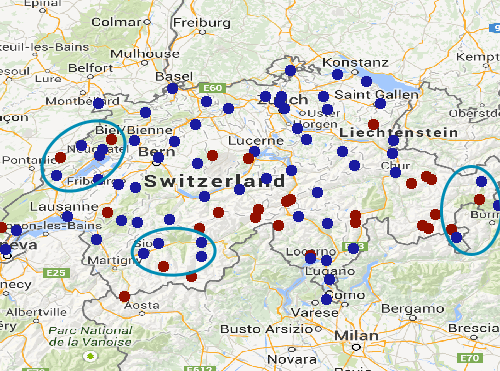}			\label{altitude_cluster_new_kmeans_labeled}}
        \vspace{-0.2cm}
        \caption{Two clusterings of the measuring stations obtained by (a) applying spectral clustering to the learned graph, and (b) applying $k$-means to the raw temperature records. The red cluster includes stations at higher altitudes and the blue cluster includes stations at lower altitudes. Some of the stations with high altitudes, highlighted by the cyan ellipses, are clustered together with low-altitude stations by $k$-means.}
        \label{fig:altitude_cluster}
\end{figure*}

To further verify our results, we separate the measuring stations into disjoint clusters based on the graph learned by \textbf{GL-SigRep}, such that different clusters correspond to different characteristics of the measure stations. In particular, since we obtain a valid Laplacian matrix, we can apply spectral clustering \cite{Ng02} to the learned graph to partition the vertex set into two disjoint clusters. The results are shown in Fig.~\ref{fig:altitude_cluster}(a), where the red and blue dots represent two different clusters of stations. As we can see, the stations in the red cluster are mainly those located on the mountains, such as those in the Jura Mountains and Alps, while the ones in the blue cluster are mainly stations in flat regions. It is especially interesting to notice that, the blue stations in the Alps region (from centre to the bottom middle of the map) mainly lie in the valleys along main roads (such as those in the canton of Valais) or in the Lugano region. 
For comparison, we show in Fig.~\ref{fig:altitude_cluster}(b) a 2-cluster partition of the stations by applying $k$-means clustering algorithm using directly the observed temperature records. We see that the clusters obtained in this way are less consistent with the altitude information of the stations. For instance, some of the stations in mountain regions in Jura, Valais and eastern Switzerland (highlighted by the cyan ellipse) have been clustered together with low-altitude stations. 
One possible explanation could be that the $k$-means algorithm treats the observations at each vertex of the graph independently and does not emphasize a global relationship during the clustering process, therefore it may not be able to capture a global similarity pattern as required in our graph learning task.
These results show that the clusters obtained by \textbf{GL-SigRep} capture the altitude information of the measuring stations hence confirms the quality of the learned graph topology.

\subsection{Learning climate graph from evapotranspiration data}
\begin{figure}[t]
	\begin{center}
		\begin{tabular}{cc}
			~\includegraphics[width=0.34\textwidth]{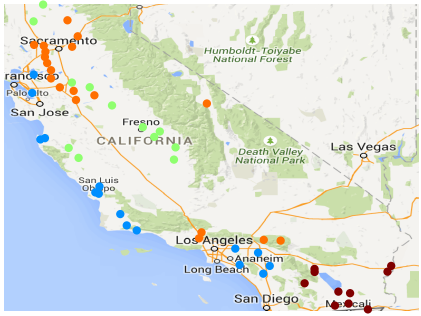}~
		\end{tabular}
	\end{center}
	\vspace{-0.2cm}
	\caption{Locations of the 56 measuring stations in California. The colors indicate 4 groundtruth clusters that correspond to 4 ETo zones.}
	\label{fig:cimis_gt_clus}
\end{figure}

In this application, we consider average monthly evapotranspiration (ETo) data recorded at 56 measuring stations in California between January 2012 and December 2014 \cite{cimis}. Similarly to the previous example, for each station, we compute the average evapotranspiration for each month over the three years. This leads to 12 signals that measure the average evapotranspiration at the 56 measuring stations in the 12 months. By applying our graph learning framework, we would like to infer a graph that captures the similarities between these measuring stations in terms of the monthly variations of evapotranspiration at their locations. In this example, we do not have a readily available groundtruth graph; however, we have a reference ETo Zone Map \cite{cimis_zonemap}, which classifies each of the 56 stations into one of the 4 ETo zones according to the terrain and climate conditions at its location\footnote{In our experiment, we consider stations from the four largest zones that contain at least 10 stations, namely, zone 6, 12, 14 and 18.}. This can thus be considered as a groundtruth clustering of the 56 stations, as shown in Fig.~\ref{fig:cimis_gt_clus}. Therefore, for performance evaluation, we apply spectral clustering to the learned graphs using \textbf{GL-SigRep} and \textbf{GL-LogDet}, respectively, and we partition the stations into 4 clusters. We then compare the resulting clusters with the groundtruth information.

In Table~\ref{tab:real_quan2}, we show the overall best scores achieved by the two algorithms in terms of three evaluation metrics for clustering, namely, \textit{NMI}, \textit{Purity} and \textit{Rand Index (RI)} \cite{Manning08}. Even though this clustering task is challenging - indeed, according to the descriptions of the ETo Zone Map, there exists only slight difference between zone 12 in green and zone 14 in orange - we nevertheless see from Table~\ref{tab:real_quan2} that the clusters obtained based on the graph learned by \textbf{GL-SigRep} is more consistent with the groundtruth than that based on \textbf{GL-LogDet}.

\begin{table}[t]
\caption{Performance for \textbf{GL-SigRep} and \textbf{GL-LogDet} in terms of recovering groudtruth clusters of measuring stations.}
\centering
\begin{tabular}{c c c c c}
\hline\hline
Algorithm& NMI & Purity & RI \\ [0.5ex] 
\hline
\textbf{GL-SigRep}& 0.5813 & 0.7321 & 0.8039 \\
\textbf{GL-LogDet}& 0.5093 & 0.7321 & 0.7844 \\[1ex]
\hline
\end{tabular}
\label{tab:real_quan2}
\vspace*{-0.5cm}
\end{table}

\subsection{Learning political graph from votation data}
\begin{figure*}[t]
	\begin{center}
		\begin{tabular}{cc}
			~\includegraphics[width=0.40\textwidth]{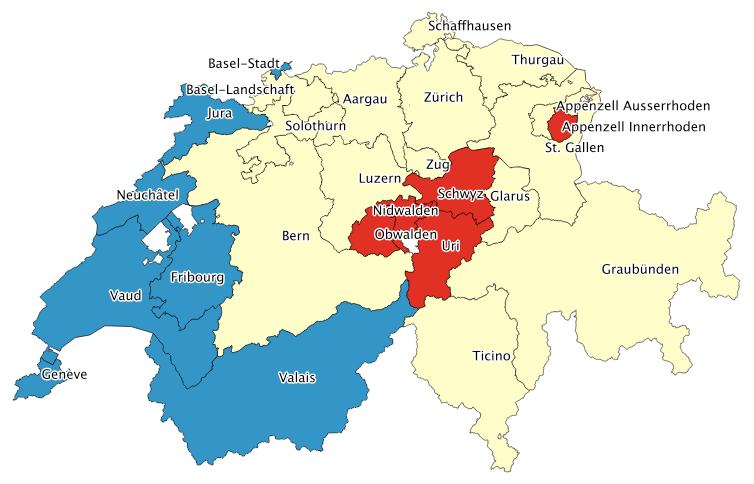}~ & ~\includegraphics[width=0.40\textwidth]{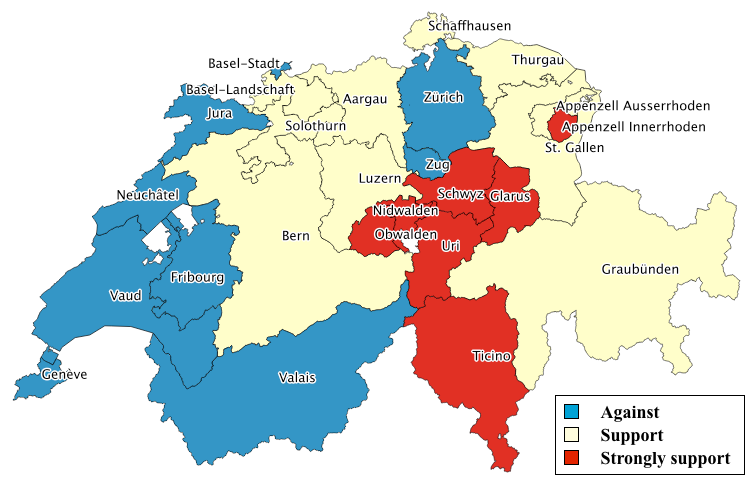}~ \\
			~(a)~ & ~(b)~
		\end{tabular}
	\end{center}
	\vspace{-0.2cm}
	\caption{(a) Clustering of the 26 Swiss cantons based on their votes in the national referendums between 2008 and 2012. (b) The results in the 26 Swiss cantons of the national referendum for the federal popular initiative ``Against mass immigration''.}
	\label{fig:voting_cluster}
\end{figure*}

We now move onto the final real world example, where we consider votation data from the national referendums for 37 federal popular initiatives in Switzerland between 2008 and 2012 \cite{swissvotes}. Specifically, we consider the percentage of votes supporting each initiative in the 26 Swiss cantons as our observed signal. This leads to 37 signals (i.e., one per initiative), each of dimension 26. By applying the proposed graph learning framework, we would like to infer a graph that captures the political relationships between the Swiss cantons in terms of their votes in the national referendums. In this example, we neither have an obvious groundtruth relationship graph nor a groundtruth partitions of the cantons in terms of their political preferences. Therefore, we focus in this example on validating our results, by interpreting the clusters obtained by partitioning the learned graph using \textbf{GL-SigRep}.

In Fig.~\ref{fig:voting_cluster}(a), we show one 3-cluster partition obtained by applying spectral clustering to the graphs learned by \textbf{GL-SigRep}. We can see that the blue cluster contains all the French-speaking cantons, while the yellow clusters contain most of the German-speaking cantons and the Italian-speaking canton Ticino. Then, the five cantons in the red cluster, namely, Uri, Schwyz, Nidwalden, Obwalden and Appenzell Innerrhoden, the first four of which constituting the so-called ``primitive'' cantons at the origin of Switzerland, are all considered among the most conservative cantons in Switzerland. The cluster membership of the canton Basel-Stadt can be explained by the fact that Basel regularly agrees with the cantons in the French-speaking part of Switzerland on referendums for close relations with the European Union. Therefore, this clustering result demonstrates that the graph learned by \textbf{GL-SigRep} indeed captures the hidden political relationships between the 26 Swiss cantons, which are consistent with the general understanding of their voting behaviors in the national referendums.

Finally, to confirm the meaningfulness of the clusterings shown in Fig.~\ref{fig:voting_cluster}(a), we illustrate in Fig.~\ref{fig:voting_cluster}(b) a 3-cluster partition of the cantons based on votation statistics in a recent national referendum for the initiative ``Against mass immigration''. This referendum, which was held in February 2014, has the largest turnout of 55.8\% in recent years. In Fig.~\ref{fig:voting_cluster}(b), the cantons in the blue cluster voted against the initiative, while the ones in the yellow and red clusters voted for it. In particular, the seven cantons where the percentage of voters supporting the initiative is greater than 58.2\% are grouped in the red cluster. As we can see, this partition is largely consistent with those in Fig.~\ref{fig:voting_cluster}(a), with the only exceptions of the cantons of Z\"{u}rich and Zug agreeing with the French-speaking cantons with small margins. This confirms the clustering in Fig.~\ref{fig:voting_cluster}(a) hence demonstrates the quality of the graphs learned by \textbf{GL-SigRep}.

\section{Conclusion}
\label{sec:conclusion}
In this paper, we have presented a framework for learning graph topologies (graph Laplacians) from signal observations under the assumption that the signals are smooth on the learned graph. Specifically, we have developed a method for learning graphs that enforces the smoothness property of the graph signals, under a Gaussian prior distribution imposed on the latent variables in a generalized factor analysis model. We believe the proposed graph learning framework can benefit numerous signal processing and learning tasks on graphs through the learning of an appropriate graph topology that captures relationships between data entities. Furthermore, it opens new perspectives in the field of graph signal processing. In particular, although we only focus on a Gaussian prior in the present paper, we can also impose other statistical priors on the latent variables in the generalized factor analysis model. 
We leave such an analysis for future work.

\section*{Acknowledgement}
The authors would like to thank Prof. Antonio Ortega and Prof. Michael Bronstein for comments on the draft, and Prof. Jean-Christophe Pesquet, Prof. Daniel Kressner, Dr. Jonas Ballani, Chenhui Hu and Vassilis Kalofolias for discussions about the optimization problem and alternative solutions.

\bibliographystyle{IEEEtran.bst}
\bibliography{mybibfile.bib}

\end{document}